\documentclass[preprint,3p]{elsarticle}

\usepackage{amssymb}
\usepackage[caption=false]{subfig}
\usepackage{graphicx} 
\usepackage{supertabular,booktabs}
\usepackage{amsmath,amssymb}   

\usepackage{algorithm}
\usepackage{algpseudocode}
\algrenewcommand\algorithmicrequire{\textbf{Input:}}
\algrenewcommand\algorithmicensure{\textbf{Output:}}

\usepackage{multirow} 
\usepackage{multicol}
\usepackage{makecell}
\usepackage{diagbox}
\usepackage[colorlinks]{hyperref}
\usepackage{longtable}
\usepackage[table,xcdraw]{xcolor}
\usepackage{booktabs}
\usepackage{threeparttable, tablefootnote}
\usepackage{lscape}
\usepackage{tabularx}

\journal{Swarm and Evolutionary Computation}

\begin{document}

\begin{frontmatter}



\title{Multiobjective Differential Evolution Enhanced with Principle Component Analysis for Constrained Optimization\tnoteref{t1}}
\tnotetext[t1]{This work was supported by the National Natural Science Foundation of China under Grant 61573157}

\author[HUANG]{Wei Huang}
\address[HUANG]{Tianjin Key Laboratory of Intelligent and Novel Software Technology, and School of Computer Science and Engineering, Tianjin University of Technology, China}
\ead{huangwabc@163.com}
\author[XU]{Tao Xu}
\address[XU]{Department of Computer Science, Aberystwyth University, Aberystwyth  SY23 3DB, UK}
\ead{tax2@aber.ac.uk}

\author[LI]{Kangshun Li}
\address[LI]{College of Mathematics and Informatics, South China Agricultural University, Guangzhou 510642, China}
 \ead{likangshun@sina.com}
 
\author[HE]{Jun He\corref{cor}}
\address[HE]{School of Science and Technology, Nottingham Trent University, Nottingham NG11 8NS, UK}
\ead{jun.he@ntu.ac.uk}
\cortext[cor]{Corresponding author}
\begin{abstract}
Multiobjective evolutionary algorithms (MOEAs) have been successfully applied  to a number of constrained optimization problems. Many of them adopt mutation and crossover operators from differential evolution. However, these operators do not explicitly  utilise features of fitness landscapes. To improve the performance of  algorithms, this paper aims at designing a search operator adapting to fitness landscapes. Through an  observation, we find that principle component analysis (PCA) can be used to characterise fitness landscapes. Based on this finding, a new search operator, called PCA-projection, is proposed. In order to verify the effectiveness of PCA-projection,  we design two algorithms enhanced with PCA-projection for solving constrained optimization problems, called PMODE and HECO-PDE, respectively. Experiments have been conducted on the IEEE CEC 2017 competition benchmark suite in constrained optimisation. PMODE and HECO-PDE are compared with  the  algorithms from the IEEE CEC 2018 competition and another recent MOEA for constrained optimisation.  Experimental results show that an algorithm enhanced with PCA-projection performs better than its corresponding opponent without this operator. Furthermore,  HECO-PDE is ranked  first on all dimensions according to the competition rules. This study reveals that  decomposition-based MOEAs,  such as HECO-PDE, are competitive with best single-objective and multiobjective evolutionary algorithms for constrained optimisation, but MOEAs based on non-dominance, such as PMODE, may not.
\end{abstract}  

\end{frontmatter}

\section{Introduction}
Optimization problems in the real world often contain different types of constraints. In mathematics, a constrained optimization problem (COP) can be formulated as
\begin{align}
	\label{equCOP}
	&\min  f(\mathbf{x}),  \quad \mathbf{x}=(x_1,\cdots,x_n) \in \mathbb{R}^n,\\
	&\textrm{subject to}\left\{ 
	\begin{array}{lll} 
	\mathbf{x} \in  \Omega=\{ \mathbf{x}\mid  L_i \leq x_i\leq U_i, \; i =1, \cdots, n\} ,\\
	 g_i(\mathbf{x})\leq0,\quad    i=1,\cdots,q,\\
		h_j(\mathbf{x})=0,\quad j=1,\cdots, r,
	\end{array}\right.
\end{align}
where $\mathbf{x} \in \Omega$ is the bounded constraint. $L_i$ and $R_i $ denote  lower and upper boundaries respectively. 
$g_i(\mathbf{x})\le 0$ is the $i$th inequality constraint. $ h_j(\mathbf{x})=0$ is the $ j $th equality constraint.

There exist a variety of evolutionary algorithms (EAs) for solving COPs, which employ different constraint handling techniques, such as the  penalty function method,  feasibility rule, repair method and multi-objective optimization~\cite{michalewicz1996evolutionary,coello2002theoretical,mezura2011constraint}. This paper focuses on the multi-objective optimization method~\cite{segura2016using}, which is to convert a COP into a multi-objective optimization problem  without any constraint. The advantage of using this method is no need to handle constraints in a special way, because constraints themselves are converted into objectives and a COP will be solved by MOEAs. 

The converted problem often is a bi-objective optimization problem~\cite{surry1997comoga,zhou2003multi}:
\begin{align} 
  \min \mathbf{f}= (f(\mathbf{x}), v(\mathbf{x})),
\end{align} 
in which one objective is the original objective function $f(\mathbf{x})$  and the other is the degree of violating the constraints $v(\mathbf{x})$.
\begin{eqnarray}
 v(\mathbf{x})=\sum_i v^g_i(\mathbf{x})+\sum_j v^h_j (\mathbf{x}).
\end{eqnarray} 
The first part in the formula is the sum of the degree of violating an inequality constraint,  given by 
\begin{eqnarray}
&v^g_{i}(\mathbf{x})= \max \{ 0, g_i(\mathbf{x}) \}, & i=1, \cdots, q.
\end{eqnarray}
The second part is the sum of the degree of violating an equal constraint, given by
\begin{eqnarray} 
&v^h_j(\mathbf{x})= \max \{0, \lvert h_j(\mathbf{x})\rvert-\delta\}, & j=1, \cdots, r,
\end{eqnarray}
where $\delta$ is  a  tolerance allowed for the equality constraint.    

Many MOEAs have been proposed for solving constrained optimisation problems~\cite{segura2016using}. Most of them adopt mutation and crossover operators from differential evolution (DE). However, these operators do not explicitly utilise characteristics of fitness landscapes. Intuitively,  a search operator which adapts to fitness landscapes may be more efficient than those without adaptation.  A recent theoretical study also claims that in terms of the time-based fitness landscape,  unimodal functions are the easiest and deceptive functions to EAs are the hardest~\cite{he2015easiest}. Therefore, it is important to design a landscape-adaptive search operator which may improve the performance of EAs.

Our idea is to enhance EAs with principle component analysis (PCA). There exist a suite of studies which have shown PCA may improve the performance of EAs.  Munteanu and Lazarescu~\cite{munteanu1999improving} designed a PCA-mutation operator and claimed that PCA-mutation is more successful in maintaining population diversity during search. 
Because of PCA's inherent capability of rebuilding a new coordinate system, Li et al.~\cite{li2012differential} applied PCA to the design of crossover for reducing correlations among variables. PCA was used  in particle swam optimization (PSO) to  mine population information for promising principal component directions~\cite{chu2011fortify,zhao2014enhanced,ong2015automatically}. This information is utilized in velocity vectors of particles. Because PCA is a powerful tool in dimensional reduction,  it helped EAs solve high dimensional optimization problems~\cite{xu2018cooperative,cui2018bat}. Besides its application in designing search operators,  local principal component analysis is used for building a regularity model in multiobjective estimation of distribution algorithms~\cite{zhang2008rm,wang2012regularity}. However, the number of references of applying PCA to EAs is still very small and it is worth making further investigations. 

In this paper, we design  a new search operator adapting to fitness landscapes with the aid of PCA. PCA is used to identify  the maximal variance direction in a population. Given a ``valley'' fitness landscape in the 3-dimensional space, we observe that the direction obtained by PCA is consistent with the valley direction. Based on this observation, we  design a new search operator, called PCA-projection. The research question of this paper is whether a MOEA enhanced with PCA-projection is able to outperform its rival without this operator or other state-of-arts EAs. To answer this question, we design two MOEAs enhanced with PCA-projection, conduct  experiments on  the IEEE CEC 2017 benchmark suit for constrained optimisation competition~\cite{cec2017online} and compare them  with  EAs from the CEC 2018 competition~\cite{cec2018online}.  

The remainder of this paper is organised as follows.
Section~\ref{section:survey} reviews MOEAs for constrained optimisation. 
Section~\ref{section:2} introduces related work in differential evolution and PCA's applications in EAs. Section~\ref{section:3} explains PCA-projection in detail. Section~\ref{sec:algorithm} designs two EAs enhanced with PCA-projection. Section~\ref{section:4} reports comparative experimental results. Section~\ref{section:5} concludes this paper.

\section{Literature Review of Multiobjective Evolutionary Algorithms for Constrained Optimisation}
\label{section:survey}
The idea of applying MOEAs to constrained optimisation has attracted researchers' interest in last two decades. Surry and Radcliff~\cite{surry1997comoga} proposed constrained optimization by multi-objective genetic algorithms. They considered a COP in a dual perspective, as a constraint satisfaction problem and   an unconstrained optimization problem.  
Coello~\cite{coello2000constraint} introduced the concept of non-dominance to handle constraints into the fitness function of a genetic algorithm. Feasible individuals are ranked higher than infeasible ones, while infeasible individuals with a lower degree of constraint violation is ranked higher than those with a higher degree. 
Zhou et al.~\cite{zhou2003multi} converts a COP to a two-objective optimization model composed of  the original objective function and the degree function violating the constraints. Then they designed a real-coded genetic algorithm based on Pareto strength and Minimal Generation Gap model. 

Most MOEAs for constrained optimisation belong to the category of MOEAs based on non-dominance or Pareto ranking.   Venkatraman and Yen~\cite{venkatraman2005generic} proposed a two-phase genetic algorithm framework for solving COPs.  In the first phase, a COP is treated as a constraint satisfaction problem. In the second phase, a COP is  treated as a bi-objective optimization problem with the simultaneous optimization of the objective function and the satisfaction of the constraints. Then the Non-Dominated Sorting Genetic Algorithm (NSGA-II) is used. 
Cai and Wang~\cite{cai2006multiobjective,wang2012combining} combined multiobjective optimization with differential evolution (CMODE) to solve COPs which is based on the two-objective model. The search is guided by infeasible solution archiving and replacement mechanism. Furthermore, they provided a dynamic hybrid framework~\cite{wang2012dynamic}, which consists of  global search and local search models. More recently, Gao and Yen et al.~\cite{gao2015dual} considered COPs as a bi-objective optimization problem, where the first objective is the reward function or actual cost to be optimized, while the second objective is the constraint violations degree.  Gao et al.~\cite{gao2015multi} proposed a reverse comparison strategy based on multi-objective dominance concept. That strategy converted the original COPs to a multi-objective problem with one constraint, and weeds out worse solutions with smaller fitness value regardless of its constraints violation.  
Li et al.~\cite{li2017many} and Zeng et al.~\cite{zeng2017general} converted a COP into a dynamic constrained
many-objective optimization problem, and considered  three  types of  MOEAs,  i.e.,  Pareto  ranking-based,  decomposition-based,  and  hype-volume  indicator-based to instantiate the framework.   

Recently, MOEAs based on objective decomposition have been applied to constrained  optimisation. Xu et al.~\cite{xu2017new} constructed several helper objective functions using the weighted sum method but with static weights. Then they employed DE for optimising optimisation subproblems.  Wang et al.~\cite{wang2018decomposition} considered the weighted sum approach with dynamical weight to decompose the problem~(\ref{equBOP}) and also applied DE to solve them.  Peng et al.~\cite{peng2018novel} adopted the Chebyshev approach in objective  decomposition with biased dynamic weights. 
 
The purpose of using MOEAs for constrained optimisation is to seek the optimal feasible solution, but not to generate a uniformly distributed Pareto front. Therefore, we guess that decomposition-based MOEAs are  more flexible than MOEAs based on non-dominance or Pareto ranking, because through biased dynamic weights, we may adjust the search direction of decomposition-based MOEAs. At the beginning, a MOEA searches different directions, but at the later stage, it focuses more on the direction towards the optimal feasible solution.

\section{Two Pieces of Related Work}
\label{section:2}
Our work is linked to classical DE~\cite{storn1997differential} and the application of  PCA in EAs~\cite{munteanu1999improving}. This section reviews them one by one. 

\subsection{Differential Evolution}
\label{secDE}
DE is a popular EA for solving continuous optimization problems~\cite{storn1997differential}. In DE,  a population $P$  is represented by   $\mu$ $n$-dimensional vectors:
\begin{align}
& P = \{ \mathbf{x}_{1}, \cdots, \mathbf{x}_{\mu} \},\\
& \mathbf{x}_{i}=( x_{i,1}, x_{i,2}, \cdots, x_{i,n})^T,  i=1, 2, \cdots, \mu,
\end{align}
where  $\mu$ is the population size. Initial individuals are  chosen randomly from $[L_i, U_i]^n$. An initial individual $\mathbf{x}=( x_1, \cdots, x_n)^T $ is generated at random as follows:
\begin{equation}
x_i= L_i+(U_i-L_i) \times rand, \quad  i =1, \cdots, n,
\end{equation}
 where $rand$ is the random number $[0,1]$.

The DE algorithm consists of three operations: mutation, crossover and selection,  which are described as follows~\cite{storn1997differential,xu2017new}. 
\begin{description}
\item [Mutation:]
for each individual  $ \mathbf{x}_{i}$ where $i=1, \cdots, \mu,$ a mutant vector 
$  \mathbf{v}_{i}=(v_{i,1},v_{i,2},\cdots,v_{i,n}) 
$  is generated by
\begin{equation}
\label{equ:DEmu}
\mathbf{v}_{i} =\mathbf{x}_{r1}+F\cdot( \mathbf{x}_{r2}- \mathbf{x}_{r3})
\end{equation} 
where individuals $ \mathbf{x}_{r1}, \mathbf{x}_{r2}, \mathbf{x}_{r3}$ are chosen from $P$ at random but mutually different. They are   also chosen to be different from $\mathbf{x}_i$.  $F$  is a real and constant factor  from $[0, 2]$ which controls the amplification of the differential variation $ \mathbf{x}_{r2}- \mathbf{x}_{r3}$.
In case $\mathbf{v}_{i}$ is out of the interval $[L_i, U_i]$, the mutation operation is repeated until $\mathbf{v}_{i}$ falls in $[L_i, U_i]$.

\item [Crossover:] in order to increase population diversity, crossover is also used in DE.  The trial vector $\mathbf{u}_{i}$ is generated by  mixing the target vector $\mathbf{x}_{i}$ with the mutant vector $\mathbf{v}_{i}$. 
\\Trial vector $
 \mathbf{u}_{i} = (u_{i,1},  u_{i,2}, \cdots,  u_{i,n}) 
$
is constructed as follows:
\begin{align}
\label{equ:DEcr}
u_{i,j}=&
\begin{cases}
{v}_{i,j},      &  \text{if } rand_j(1,0)\leq CR   \text{ or } j=j_{rand}, \\
{x}_{i,j},  &   \text{ otherwise}, 
\end{cases}  
& j=1, \cdots, n, 
\end{align}
where $rand_j(0,1) $ is a uniform random number   from $ [0, 1] $. Index $ j_{rand} $ is   randomly chosen   from $ \{1, \cdots, n \}$.   $CR\in[0,1] $ denotes the crossover   constant which has to be determined
by the user. In addition, the condition ``$ j=j_{rand} $'' is used to ensure the trial vector $ \mathbf{u}_{i} $ gets at least one parameter from vector $ \mathbf{v}_{i} $.

\item [Selection:]
a greedy criterion is used to  decide whether the offspring generated by mutation and crossover should replace its parent. Trail vector $ \mathbf{u}_{i} $ is compared to target vector $ \mathbf{x}_{i} $, then the better one will be reserved to the next generation.
\end{description}

Notice that both mutation and crossover operators do not explicitly utilise features of fitness landscapes.

There exist several variants of DE algorithms. A classical DE algorithm is the   DE/Rand/1/bin DE~\cite{storn1999system} which is illustrated below.
 \begin{algorithmic}[1]	
	\State initialize a population  $P=\{\mathbf{x}_1, \cdots, \mathbf{x}_\mu \}$; 
    \State calculate fitness values of each individual in $P$;
    \While{the terminal condition is not satisfied}
    \For{$i=1, \cdots,  \mu$} 
    \State randomly select three individuals  $ \mathbf{x}_{r1}, \mathbf{x}_{r2}, \mathbf{x}_{r3}$  from $P$ at random, such that $r_1 \ne r_2 \ne r_3 \ne i$;
    \State implement mutation and crossover and generate a child $\mathbf{u}_i$ of $\mathbf{x}_i$;
    \State calculate fitness value $f(\mathbf{u}_i)$;
    \If{$f(\mathbf{u}_i) \le f(\mathbf{x}_i)$}
    \State $\mathbf{x}_i \leftarrow \mathbf{u}_i$;
    \EndIf
    \EndFor 
    \EndWhile 
\end{algorithmic}

\subsection{Application of Principle Component Analysis in Evolutionary algorithms}
It is an interesting idea to apply PCA to the  design of EAs but so far only a few research papers can be found on this topic.  Munteanu and Lazarescu's work~\cite{munteanu1999improving} designed a mutation operator based on PCA. They claimed that a PCA-mutation genetic algorithm (GA) is more successful in maintaining population diversity during search.  Their experimental results show that  a GA with the PCA-mutation obtained better solutions compared to solutions found using GAs with classical mutation operators for a filter design problem. 

Munteanu and Lazarescu~\cite{munteanu1999improving} designed a new mutation  operator on a projection search space generated by PCA, rather than the original space.  Their PCA mutation   is described as follows. A population with $\mu$ individuals is represented by an $n \times \mu$ matrix $\mathbf{X}=[\mathbf{x}_1, \cdots, \mathbf{x}_\mu ]$ where $n$ is the space dimension and $\mu$ the population size. Each $\mathbf{x}$ is an individual represented by  a column vector.  

\begin{algorithmic}[1]	
\State From the data set $\mathbf{X}$, calculate the $n\times n$ covariance matrix $\mathbf{\Sigma}$: 
\begin{align}
\mathbf{\Sigma} = \mathbb{E}[(\mathbf{x}-\mathbf{m})(\mathbf{x}-\mathbf{m})^T] 
\end{align} 
where $\mathbf{m}=\mathbb{E}[\mathbf{x}]$ which is the mean over  $\mathbf{x}_1, \cdots, \mathbf{x}_\mu$.
		
\State Given the co-variance matrix $\mathbf{\Sigma}$, compute its   eigenvectors $\mathbf{v}_1, \cdots, \mathbf{v}_n$ and  sort them in the order of the corresponding eigenvalues of these eigenvectors from high to low.  Form a $n\times n$ matrix $\mathbf{V}=[\mathbf{v}_1, \cdots, \mathbf{v}_n]$.

\State Calculate the projection of the data set $\mathbf{X}$ using the
orthogonal basis  $\mathbf{v}_1, \cdots, \mathbf{v}_n$ and obtain a projected population, represented by the matrix $\mathbf{Y}= [\mathbf{y}_1, \cdots, \mathbf{y}_n]^T$: 
\begin{equation}
	\mathbf{y}_i=  \mathbf{V}^T (\mathbf{x}_i -\mathbf{m}).
\end{equation}

\State Compute the squared length of the projections along each direction $\mathbf{v}_i$, that is,
\begin{align}
\parallel L_{i}(x_j) \parallel^2 =y^2_{i,j},\quad  i =1, \cdots, n, \quad j =1, \cdots, \mu.
\end{align}

\State Choose quantities $c_{i,j}$ randomly between $0$ and $c_{\max}$ where $c_{\max}$ is a constant parameter of the mutation operator such that $c_{i-1,j}  \le c_{i,j}$ for $i=1, \cdots, n$.

\State The mutation operator adds the quantities $c_{i,j}$ to each projected squared coordinate as follows:
\begin{align}
\label{equ_PCA_mutation}
\parallel L'_i(x_j) \parallel^2  = \parallel L_i(x_j) \parallel^2 +c_{i,j}.
\end{align}

\State Compute the sign of each element in the matrix $\mathbf{Y}$, which is represented by  the matrix $signum(\mathbf{Y})$. 

\State Generate the child $\mathbf{y}'_i$ from $\mathbf{y}_i$ as follows:  $y'_{i,j}$  equals to the square roots of the mutated square projections $\parallel L'_{i} (x_j )\parallel^2$ multiplied by the corresponding sign
$signum( y_{i,j})$.

\State Obtain the mutated point in the original search space:
\begin{align}
\mathbf{x}'_i =\mathbf{V} \mathbf{y}'_i + \mathbf{m}.
\end{align} 
\end{algorithmic}

Notice that the above PCA-mutation doesn't reduce the data set $\mathbf{X}$ into a lower dimension space,  instead $\mathbf{X}$ and $\mathbf{Y}$  have the same dimension. This PCA-mutation aims to conduct mutation in the projection space rather than the original space.  However the dimensions of the projection space and original space are the same.

\section{A New Search Operator: PCA-projection} 
\label{section:3}
In order to improve the performance of EAs, we propose a new search operator, called PCA-projection, which is able to  adapt to fitness landscapes.  

\subsection{Principle Component Analysis and Valley Direction}
Although  PCA-mutation proposed in \cite{munteanu1999improving} was efficient for  a filter design problem, it  has a disadvantage.  PCA-mutation still acts on the same dimension space as the original search space. Thus, as the population size increases, the calculation of eigenvalues and eigenvectors in PCA becomes more and more expensive. In this paper, we propose a different PCA-search operator in which PCA is only applied to several selected points. A question is how to select points from a population for implementing  PCA? The solution relies on the ``valley'' concept. 

In the 3-dimensional space, a valley is intuitive which means a low area between two hills or mountains. However, this definition is really fuzzy in high-dimensional spaces.  What does a valley in a higher dimensional space mean? How to identify the location of a valley? So far there   exist no clear mathematical definition about the valley.  In this paper,  we study  the valley landscape using PCA and find that PCA provides a statistic method of identifying the valley direction. 

\begin{figure}[ht]
\begin{center}
\subfloat[]{\includegraphics[width=.45\textwidth]{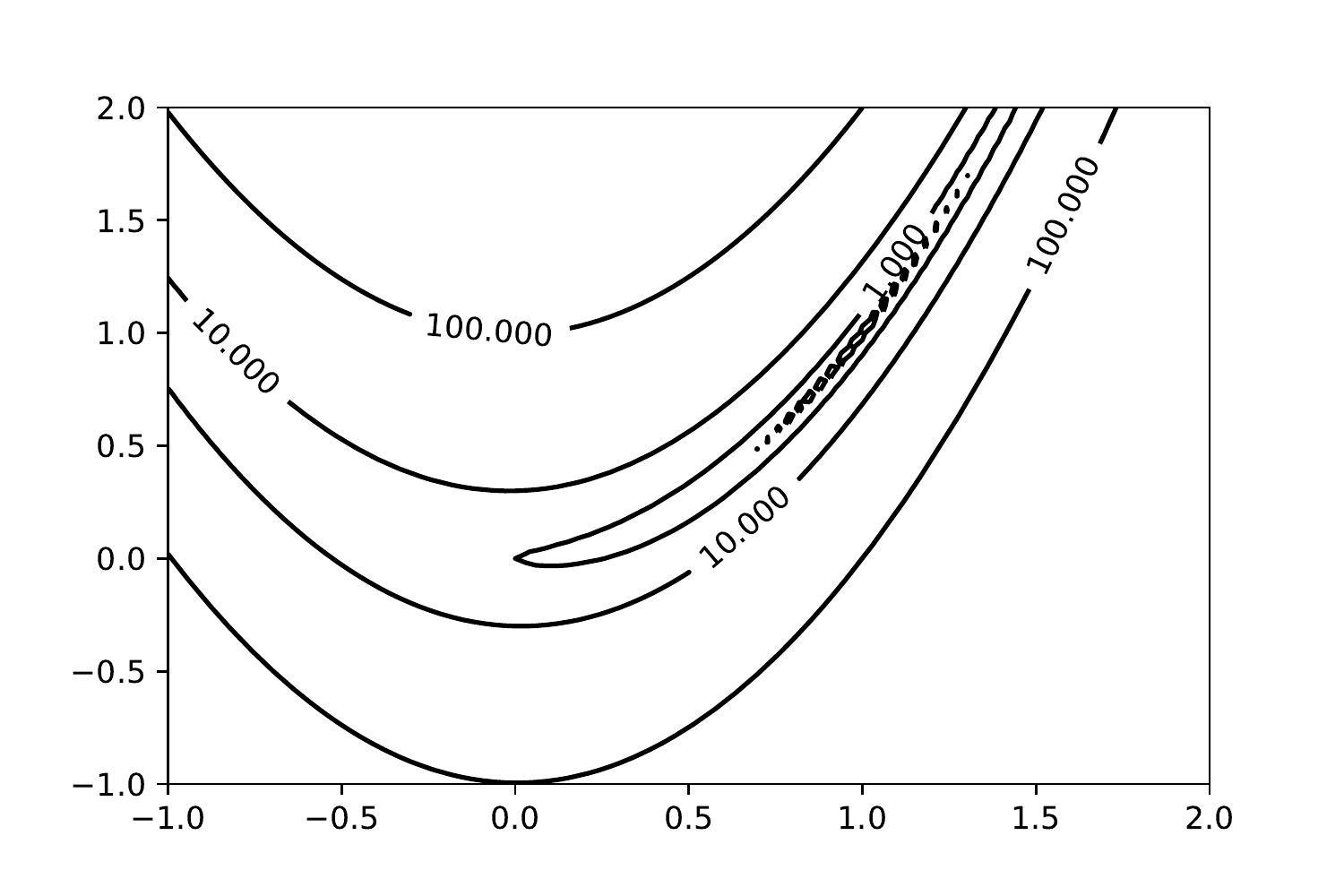}\label{figValleyA}} 
\subfloat[]{\includegraphics[width = .45\textwidth]{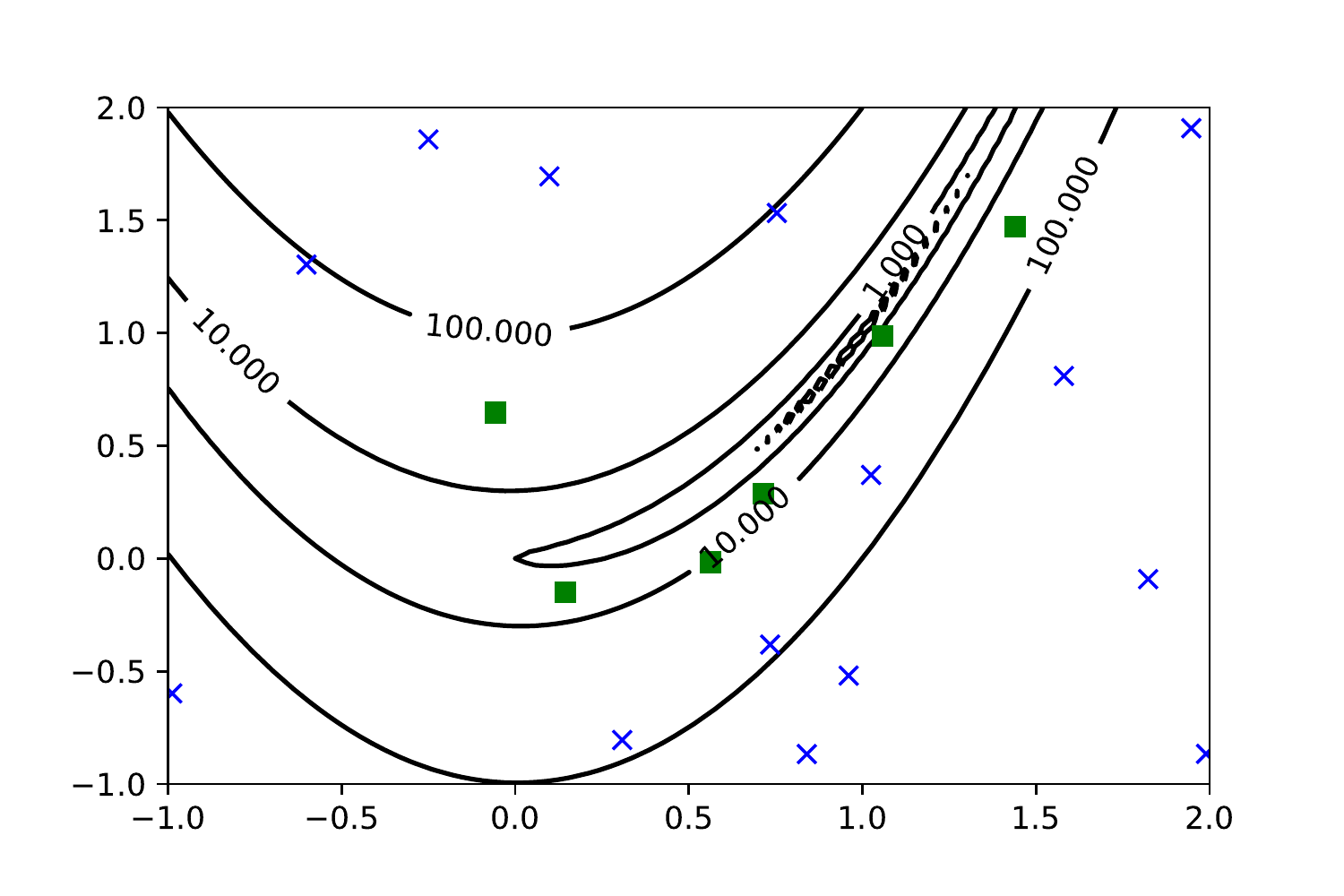}\label{figValleyB}}\\ 
\subfloat[]{\includegraphics[width=.45\textwidth]{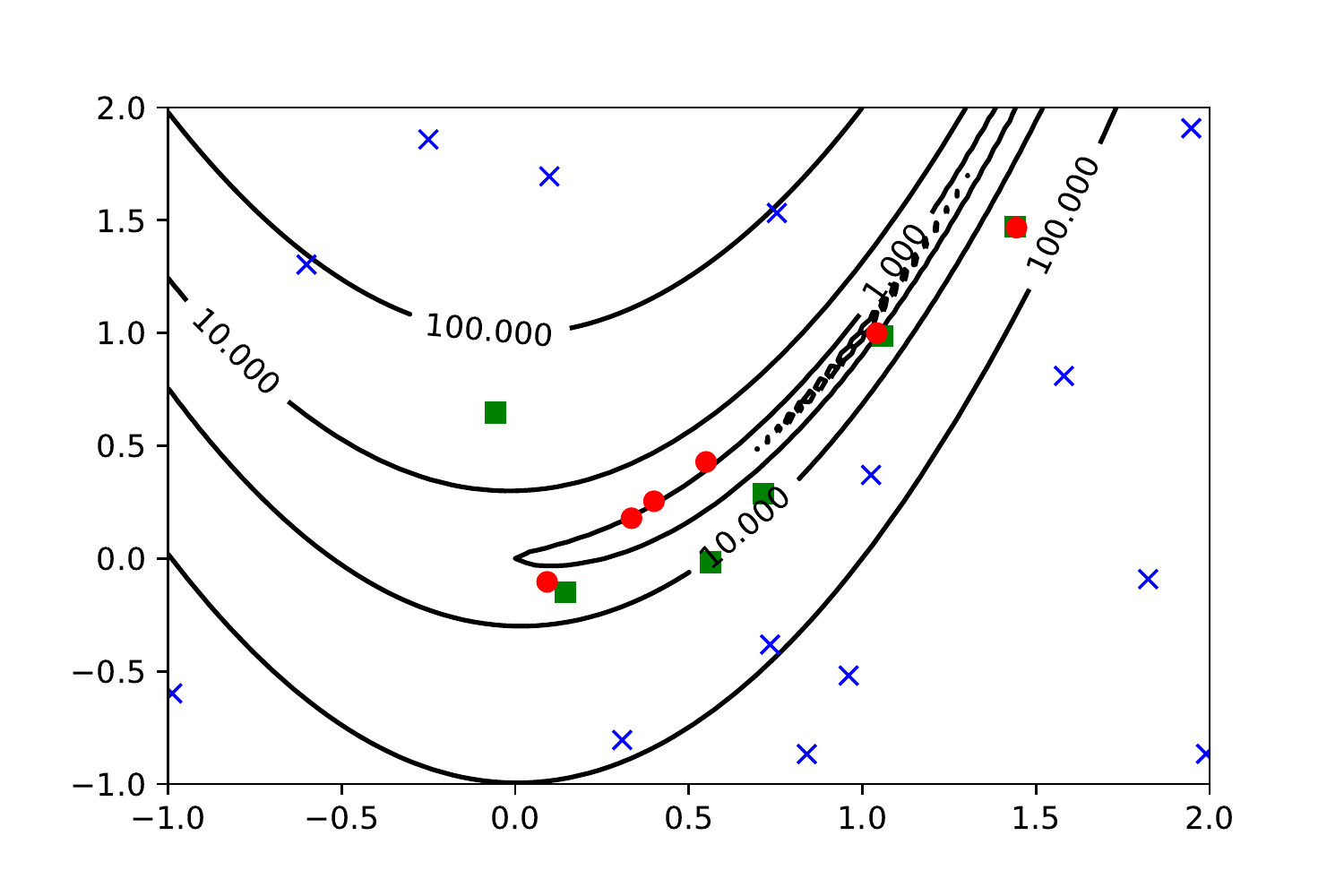}\label{figValleyC}} 
\subfloat[]{\includegraphics[width = .45\textwidth]{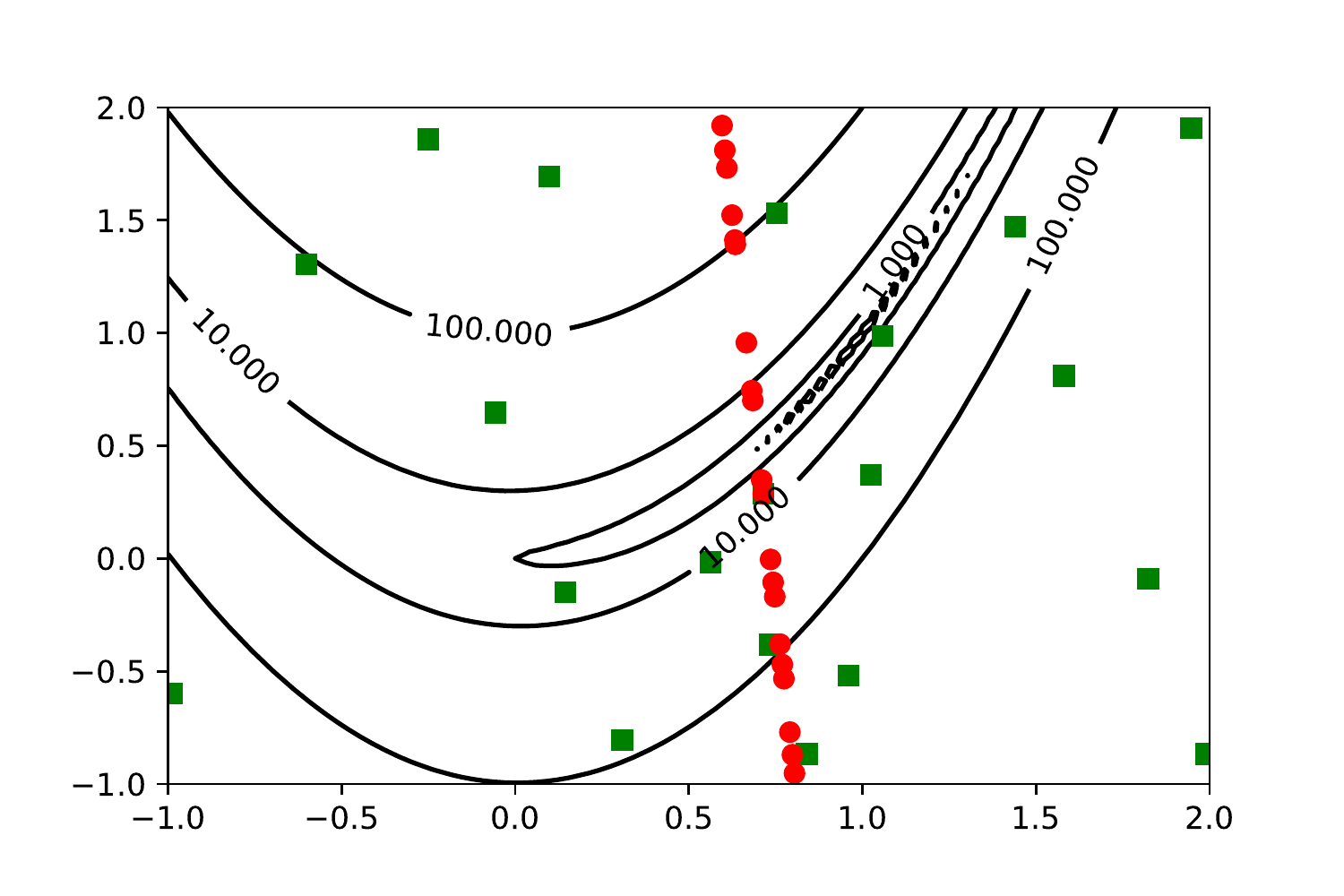}\label{figValleyD}}\\ 
\caption{PCA and the valley landscape}
\end{center}
\end{figure}

Let's explain our idea using the well-known Rosenbrock function:
\begin{equation}
f(x,y) = (1-x)^2 + 100(y-x^2), \quad - 1< x <2,-1<y<2
\end{equation}
Its minimum point  is at $(1,1)$ with $f(1,1)=0$. Fig.~\ref{figValleyA}  shows the contour graph of Rosenbrock function. From Fig.~\ref{figValleyA}, it is obvious that a deep valley exists on this landscape. But how to identify the valley?  

In the following we show a statistical method of calculating the valley direction. First we sample 20 points at random and  select 6 points with smallest function values from the population. Fig.~\ref{figValleyB}   depicts that these 6 points  (labelled by squared points) are  closer to the valley than other points.  

Next we  identify the valley direction. Since the selected 6 points distribute  along the valley,   the valley direction can be regarded as a  direction along which the variance of the 6 points is maximal. This direction can be identified by PCA. Assume that the valley direction is a linear line, the valley in fact can be approximated by the first principle component found by PCA. Let's project the 6 selected points onto the first principle component.  Fig. \ref{figValleyC}  shows that the projected points (labeled by dotted points) approximately represent the valley direction.

But  it should be pointed out if  we apply PCA to the whole population and project all 20 points onto the first principle component, we cannot obtain the valley direction.  Fig.~\ref{figValleyD}  shows that the mapped points  (labeled by dotted points)) don't distribute along the valley direction. The mapped points could represent any direction if these 20 points are sampled at random.

\subsection{Proposed PCA Projection}
Based on the observation in the above subsection, we propose a new search operator.  Here is our idea: Given a population, we select a group of points with smaller function values from the population and apply PCA to calculate  principle components; then project the points onto the principle components;  at the end, reconstruct the projected points in  the original search space. These points are taken as the children. The procedure is described in detail as follows:
\begin{description}
\label{dep:PCA-prjection}
\item 
 [PCA-projection:] Given a population $P$ and a fitness function $f(\mathbf{x})$,
\begin{algorithmic}[1]  
\State    Select $k$ individuals $\{ \mathbf{x}_1, \cdots, \mathbf{x}_k\}$  with smaller fitness values from the population $P$ ($k$ is a small constant). Denote these individuals by $\mathbf{X}$. 
 
\State Calculate the $n\times 1$ mean vector $\mathbf{m}$ and  $n\times n$ covariance matrix $\mathbf{\Sigma}$:
\begin{align}
\mathbf{m} =\frac{1}{k} \sum^k_{i=1} \mathbf{x}_i, && \mathbf{\Sigma}=\frac{1}{k-1} \sum^k_{i=1} (\mathbf{x}_i-\mathbf{m})(\mathbf{x}_i-\mathbf{m})^T.
\end{align}

\State Calculate the eigenvectors $\mathbf{v}_1, \cdots, \mathbf{v}_n$ of the covariance matrix 
$\mathbf{\Sigma}$, sorted them so that the eigenvalues of $\mathbf{v}_i$ is larger than
$\mathbf{v}_j$ for $i <j$. Select the first $m$ eigenvectors $\mathbf{v}_1, \cdots, \mathbf{v}_m$ and form a $n\times m$ matrix $\mathbf{V}=[\mathbf{e}_1, \cdots, \mathbf{e}_m]$ where $m$ is a small constant (e.g., $m=5$).  
 
\State Project $\mathbf{x}_i$ onto  the $m$-dimensional space:
\begin{align}
\mathbf{y}_i = \mathbf{V}^T (\mathbf{x}-\mathbf{m}).
\end{align} 

\State Reconstruct the projected point $\mathbf{x}_i$ in the original space: 
\begin{align}
\mathbf{x}'_i= \mathbf{m}+ \mathbf{V} \mathbf{y}_i.
\end{align} 
\end{algorithmic}
\end{description}

We call the search operator   PCA-projection, rather than PCA-mutation   \cite{munteanu1999improving}, because it does not include a mutation step.

\subsection{Characteristics of PCA-projection}
PCA-projection is a nonlinear mapping from  points in $\mathbb{R}^{n}$ to points in $\mathbb{R}^m$ which are assigned along the first $m$ principle components. It can be regarded as a multi-parent recombination operator. Like other recombination operators, it works only if the parent population keeps a degree of diversity. Otherwise, it might degenerate. For example, if  $k$ points are identical (say $\mathbf{x}$), then after PCA-projection, the projected points are still $\mathbf{x}$. If  
$k$ points distribute on the same line (say $\mathbf{y} =\mathbf{a} \mathbf{x}+\mathbf{b}$), then the first principle component is $\mathbf{y} = \mathbf{a} \mathbf{x}+\mathbf{b}$. After PCA-projection, there is no change on these $k$ points. 

PCA-projection generates the direction along which the distribution of points has the maximal variance. It is not the gradient direction. Let's show the difference through a simple example. Consider the minimisation problem $$\min f(x,y)= |x|+0\times y.$$ 
If  $k$ points  distribute on the same line (say $x_i=1, y_i=i,$ where $i=1, \cdots, k$), then the first component direction found by PCA-projection is   $x=1$.   But the gradient direction is  perpendicular to the line $x=1$.

Compared with   PCA-mutation  in~\cite{munteanu1999improving},  PCA-projection has three new characteristics:
 
\begin{enumerate}
     \item The time complexity of  PCA-projection is less than   PCA-mutation in~\cite{munteanu1999improving}. Given $k$ points in $\mathbb{R}^n$, the covariance matrix computation in PCA is $O(k^2 n)$ and its eigenvalue decomposition is $O(n^3)$. So, the complexity of PCA is $O(k^2 n+n^3)$. In PCA-projection,  only $k$ good points are sampled from the population. For example, $k=8$ in this paper, so, its time complexity is $O(8^2n+n^3)$. But in PCA-mutation, the number of points equals to the population size. For example, population size is $12n$, so, its time complexity is $O(144 n^3+n^3)$.  
     
     \item The PCA-projection has an intuitive explanation. Given a valley landscape, it projects an individual to a new position along the valley direction. 
    \item It also takes the advantage of compressing a higher dimensional data into a lower dimension space. It projects a point into a lower dimensional space.

\end{enumerate}

\section{Two New Algorithms Enhanced by PCA-Projection}
\label{sec:algorithm}
In this section, we present two MOEAs enhanced with PCA-projection for COPs.

\subsection{New Algorithm 1: PMODE = CMODE + PCA-projection} 
\label{subsection:CMODE}
We design the first algorithm though adding PCA-projection to CMODE, an algorithm proposed by Cai and Wang~\cite{cai2006multiobjective,wang2012combining}. It combines multiobjective optimization with differential evolution. CMODE is used to solve the standard bi-objective problem: 
\begin{eqnarray} 
\label{equBOP}
  \min (f(\mathbf{x}), v(\mathbf{x})), \quad x \in \Omega.
\end{eqnarray} 

CMODE belongs to the family of MOEAs based on non-dominance. At each generation, it identifies non-dominant solutions under functions $(f,v)$ and replaces those dominated solutions. 
It is straightforward  to add PCA-projection into CMODE through the mixed strategy, that is to apply PCA-projection with probability $p$ and normal mutation and crossover operations with probability $1-p$. 
After  PCA-projection is added into CMODE, we name the new algorithm PMODE. The pseudo-code of  PMODE is shown as below.   

\begin{algorithmic}[1] 
\State initialise  a population $P$  of solutions at random;
\label{stepI1}
\State evaluate  the fitness value $f$ and constraint violation $v$  for each individual in $P$;  
\label{stepI2}
\State set $FES=\mu$; 
\label{stepI3}
\State set $A=\emptyset$; 
\label{stepI4}
  
\For{$t=1, \cdots, FES_{\max}$} 
\label{stepFor}
\State initialise children population $C = \emptyset$;
\label{stepInitialiseC}
\State  randomly choose $\lambda$ individuals (denoted by $Q$) from population $P$;
\label{stepP1}        	 
\State let $P'=P \setminus Q$;
\label{stepP2}

\For{$\mathbf{x}_{i}$ in $Q$, $i = 1,\dots,\lambda$}
\If{$rand < 0.1$}
\label{stepIf1}
 \State  perform PCA-projection on set $Q$ and generate $\lambda$ projected solutions $\mathbf{x}'_1, \cdots, x'_{\lambda}$;
\label{stepPCA1}
\State let $\mathbf{y}_{i} = \mathbf{x}'_i$; 
\label{stepPCA2}
\Else 
\State generate a trail vector $\mathbf{y}_{i}$ by mutation and crossover;
\label{stepMutationCrossover}
\EndIf
\label{stepEndIf1}
\State update $C \leftarrow C \cup \{\mathbf{y}_{i}\}$;        	\label{stepChildren}
\EndFor
\State evaluate the  values of $f$ and  $v$   for each individual in   $C$;  
     	 and update $FES  \leftarrow FES+\lambda$; \label{stepEvaluate2} 
\State identify all non-dominated individuals  in  $C$ (denoted by $R$);
\label{stepNondominated}
\For{each individual $\mathbf{x}$ in $R$}\label{stepSelectionStart} 
\State find all individual(s) in $Q$ dominated by $\mathbf{x}$; 
\label{stepReplace1}
\State randomly replace one of these dominated individuals by $\mathbf{x}$;
\label{stepReplace2}
\EndFor 
\label{stepSelectionEnd}
\State let $P=P' \cup Q$;
\label{stepNewPopulation}
 \If{no feasible solution exists in $R$}
 \label{stepNoFeasible}
 \State  identify the infeasible solution $\mathbf{x}$ in $R$ with the lowest degree of constraint violation and add it to archive $A$;
 \EndIf
 \If{$\mod (t,k)=0$} 
 \State execute the infeasible solution replacement mechanism and set $A=\emptyset$; 
 \EndIf
 \label{stepNoFeasible2}
       	\EndFor
       	\label{stepEndFor}
 	    \Ensure the best found solution  
    \end{algorithmic}

For the sake of self-contained presentation, we explain PMODE in detail. Lines~\ref{stepI1}-\ref{stepI4} are initialisation steps. In lines~\ref{stepI1} and \ref{stepI2},  an initial population $P$ is generated at random, in which all initial vectors are  chosen randomly from $[L_i, U_i]^n$. Then the fitness value  and constraint violation degree of each individual are evaluated. In line~\ref{stepI3}, $FES$ records the number of fitness evaluations. In line~\ref{stepI4},  $A$ is an archive to store the infeasible individual with the lowest degree of constraint violation. Initially it is set to an empty set. 

Lines~\ref{stepFor} to \ref{stepEndFor} are the main loop of this algorithm.  $FES_{\max}$  represents the maximum number of function evaluations. In line~\ref{stepInitialiseC},  $C$ is a set  to keep the children generated by PCA-
projection, mutation and crossover. Initially it is set to an empty set. 

In lines~\ref{stepP1} and \ref{stepP2}, $\lambda$ individuals (denoted by $Q$) are selected from population $P$.   $P'$ denotes the rest individuals in $P$. Subpopulation $Q$ takes part in PCA-projection, mutation and crossover, but subpopulation $P'$  does not.

In  lines~\ref{stepIf1} to~\ref{stepEndIf1}, a trial vector $y$ is generated by a mixed strategy.   PCA-projection is chosen with probability 0.1, or mutation and crossover are chosen with probability 0.9. $rand$ is a random number in $[0,1]$. This is the main difference form CMODE~\cite{wang2012combining} in which PCA-projection exists.  

In line~\ref{stepPCA1},  PCA-projection is conducted on subpopulation $Q$, then $\lambda$ new solutions are generated. The procedure of PCA-projection on $Q$ is given as follows: (1) calculate the $5$ eigenvectors $\mathbf{v}_1, \cdots, \mathbf{v}_n$ of the covariance matrix with the largest eigenvalues; (2) project $\mathbf{x}_i$ onto  a lower-dimensional space; (3) reconstruct the projected point $\mathbf{x}_i$ in the original space, denoted by $x'_i$.  In line~\ref{stepPCA2}, set the trial vector $\mathbf{y}_{i}=\mathbf{x}'_i.$

In line~\ref{stepMutationCrossover},  generate a mutant vector 
$  \mathbf{v}_{i}$  by
\begin{equation}
\mathbf{v}_{i} =\mathbf{x}_{r1}+F \cdot( \mathbf{x}_{r2}- \mathbf{x}_{r3})
\end{equation} 
where random indexes $ r1, r2, r3$ are mutually different  integers. They are   also chosen to be different
from the running index $i$.  $F$  is a real and constant factor  from $[0, 2]$ which controls the amplification of the differential variation $( \mathbf{x}_{r2}- \mathbf{x}_{r3})$. 

Then the trial vector $\mathbf{y}_{i}$ is generated by crossover on the mutant vector $\mathbf{v_i}$. In more detail, trial vector $\mathbf{y}_{i} = (\mathbf{y}_{i,1}, \cdots,  \mathbf{y}_{i,n}) $ is constructed as follows:
 \begin{align} 
y_{i,j}=\left\{
\begin{array}{ll}
\mathbf{v}_{i,j},      &  \text{if } rand_j(1,0)\leq CR    \text{ or } j=j_{rand}, \\
 \mathbf{x}_{i,j},  &   \text{ otherwise}, 
\end{array}
\right.  && j=1, \cdots, n, 
\end{align}
where $rand_j(0,1) $ is a uniform random number   from $ [0, 1] $. Index $ j_{rand} $ is   randomly chosen   from $ \{1, \cdots, n \}$.   $CR\in[0,1] $ denotes the crossover   constant which has to be determined
by the user. In addition, the condition ``$j=j_{rand} $'' is used to ensure the trial vector $ \mathbf{y}_{i} $ gets at least one parameter from vector $ \mathbf{v_i} $.

In line~\ref{stepChildren}, the trail vector $\mathbf{y}_i$ is added into children subpopulation $C$.

In line~\ref{stepEvaluate2},  the values of $f$ and $v$ of each individual in $C$ are evaluated. Then in line~\ref{stepNondominated}, all non-dominated solutions in $C$ are identified and stored in set $R$. In lines~\ref{stepReplace1} and~\ref{stepReplace2}, non-dominated individuals  in  $R$ replace dominated individuals in  $Q$ at random. 

In line~\ref{stepNewPopulation}, a new generation population $P$ is generated by merging $P' \cup Q$.  

Lines~\ref{stepNoFeasible} to~\ref{stepNoFeasible2} are the infeasible solution replacement mechanism, which is the same as CMODE~\cite{wang2012combining}. 
 
The difference between PMODE and CMODE is PCA-projection. There is no other change in other parts. This design aims to evaluate the effectiveness of PCA-project without the interface of other factors.

\subsection{New Algorithm 2: HECO-PDE = HECO-DE +PCA-projection}
We design the second algorithm through adding PCA-projection to HECO-DE\footnote{Xu, T., He, J.,\& Shang, C. (2019). Helper and Equivalent Objectives: An Efficient Approach to Constrained Optimisation. arXiv preprint arXiv:1903.04886. Also in \textit{ACM GECCO 2019 Competition in Constrained Optimisation}.}, an algorithm proposed by Xu, He and Shang under the framework of  helper and equivalent objectives. HECO-DE is built upon a single-objective EA (LSHADE44) for constrained optimisation~\cite{polakova2017shade}, but  aims at minimising two objective functions within  population $P$: 
\begin{align}
\label{equHEOP}
\min \mathbf{f}(\mathbf{x}) =(e(\mathbf{x}),   f(\mathbf{x})), 
\quad 
 \mathbf{x} \in P. 
\end{align}
This bi-objective optimisation problem is different from the standard one (\ref{equBOP}) because function $e(\mathbf{x})$ is given by 
\begin{equation}
 \label{equEquivalentFunction}
    e(\mathbf{x})= 
    w_1 \tilde{e}(\mathbf{x}) +w_2 v(\mathbf{x}), 
\end{equation} 
where $w_1, w_2\ge 0$ are weights.   Function $\tilde{e}(\mathbf{x}) =|f(\mathbf{x})-f^*(P)|$
in which $f^*(P)$ denotes the best individual's fitness in population $P$, given as  
\begin{align*}
    f^*(P)=
    \left\{
    \begin{array}{ll}
     \min\{v(\mathbf{x}); \mathbf{x} \in P\},      & \mbox{if }   P\cap \Omega_F=\emptyset,  \\
    \min\{f(\mathbf{x}); \mathbf{x} \in P\cap \Omega_F\},      & \mbox{if }   P \cap \Omega_F\neq \emptyset.
    \end{array}
    \right.
\end{align*}

HECO-DE belongs to the family of MOEAs based on decomposition. Problem~(\ref{equHEOP}) is decomposed into $\lambda$  single objective subproblems through the weighted sum method: for $ 
   i=1, \cdots, \lambda,$ 
\begin{align}
\label{equHEOP4}
\begin{array}{rr}
     &   \min f_i(\mathbf{x})=  w_{1i} \tilde{e}(\mathbf{x}) + w_{2i} v(\mathbf{x})+w_{3i} f(\mathbf{x}), 
\end{array}
\end{align} 
where $w_{1i}, w_{2i}, w_{3i} \ge 0$ are $\lambda$ tuples of weights. 

An extra term $\tilde{e}$ is added besides the original objective function $f$ and constraint violation degree $v$, so  problem (\ref{equHEOP4}) is equivalent to a tri-objective optimisation problem.
\begin{align}
\label{equHEOP5}
\min \mathbf{f}(\mathbf{x}) =(\tilde{e}(\mathbf{x}), f(\mathbf{x}),   v(\mathbf{x})), 
\qquad
 \mathbf{x} \in P. 
\end{align}

PCA-projection is added into HECO-DE through a mixed strategy, that is to apply PCA-projection with probability $p$ and normal mutation and crossover operations with probability $1-p$. For the new algorithm enhanced with PCA-projection, we call it HECO-PDE. The pseudo-code of HECO-PDE is shown as follows.

\begin{algorithmic}[1]
    \Require objective function $f(\mathbf{x})$ and constraint violation  $v(\mathbf{x})$; 
    \State initialise algorithm parameters: the required sizes $\mu_0$, $\mu_{T_{\max}}$; circle memories $M_{F}$ and $M_{CR}$, memory size $H$; probabilities $q_k = 1/4,k =1,\cdots,4$, archive $A = \emptyset$; 
        \label{HECO-Initialisation1}
    \State   $t = 0$;
        \label{HECO-Initialisation2}
    \State randomly generate an initial population $P$ of size $\mu_0$;
        \label{HECO-Generate-Population}
    \State evaluate   $f(\mathbf{x})$  and  $v(\mathbf{x})$  for  $\mathbf{x} \in P$;
    \label{HECO-Evaluate1}
    \State $FES = \mu_0$; 
    \label{HECO-FES}
    \While {$FES \le FES_{\max}$ (or $t \le T_{\max}$)}
    \label{HECO-While-Start}
        \State {adjust weights;}
        \label{HECO-Weight}
        \State for each strategy, set $S_{F} =\emptyset ,S_{{CR}}=\emptyset$; 
        \label{HECO-Stratgies}
        \State children population $C=\emptyset$;
        \label{HECO-Children}
        \State {randomly select $\lambda$ individuals (denoted by $Q$) from $P$ and then set $P'= P \setminus Q$;}
        \label{HECO-population-Q}
        \For{$x_i$ in $Q$, $i = 1,\dots,\lambda$}
        \label{HECO-For-Start}
           \If{$rand < 0.1$}
           \label{HECO-If-Start}
               \State perform  PCA-projection on set $Q$, and generate $\lambda$ projected individuals $(\mathbf{x}'_{1}, \cdots, x'_{\lambda}$);
               \label{HECO-PCA1}
        	        \State let trail vector $\mathbf{y}_{i} = \mathbf{x}_{i}'$; 
        	        \label{HECO-PCA2}
        	    \Else 
        	     \State choose one strategy  subject to  probabilities $q_k, k=1, \cdots, 4$ and generate parameters $F$ and $CR$ with respective circle memories;
            \label{HECO-Stratgies2}
        	        \State{generate trail vector $\mathbf{y}_{i}$ by mutation and crossover;}
        	        \label{HECO-trail}
        	   \EndIf
        	   \label{HECO-If-End}
            \State {evaluate $f(\mathbf{y}_{i})$ and $v(\mathbf{y}_{i})$;}
            \State {$Q' \leftarrow Q \cup \{\mathbf{y}_{i}\}$;}
            \label{HECO-append}
            \State normalise  $\tilde{e}(\mathbf{x})$, $f(\mathbf{x})$ and $v(\mathbf{x})$ in $Q'$;
            \label{HECO-normalize}
            \State{calculate $f_i(\mathbf{y}_{i})$ and $ f_i(\mathbf{x}_{i})$;}
            \label{HECO-calculate-fi-vi}
            \If {$f_i(\mathbf{y}_{i}) < f_i(\mathbf{x}_{i})$}
               \State {update children population  $C \leftarrow \mathbf{y}_i \cup C$ and archive  $A \leftarrow \mathbf{x}_i \cup A$;}
               \label{HECO-update-archive}
                \State {save the values of $F$ and $CR$ into respective sets $S_{F}$ and $S_{CR}$ and increase respective success count;}
                \label{HECO-save-parameters}
            \EndIf
        \EndFor
        \label{HECO-For-End}
        \State {update circle memories $M_{F}$ and $M_{CR}$  using respective sets $S_F$ and $S_{CR}$ for each strategy;}
        \label{HECO-update-memory}
        \State {$P \leftarrow P' \cup C$;}
        \label{HECO-update-P}
        \State {calculate the required population size $\mu_{t}$;}
        \label{HECO-population-reduce1}
        \If {$\mu_{t} < |P|$}
            \State{randomly remove $|P|-\mu_{t} $ selected individuals from $P$;}
        \EndIf
        \label{HECO-population-reduce2}
         \State calculate $|A|_{\max} =4 \mu_{t}$; 
         \label{HECO-archive1}
        \If {$|A| > |A|_{\max}$ }
            \State{randomly delete $|A| - |A|_{\max}$ selected individuals from archive $A$};
        \EndIf
        \label{HECO-archive2}
        \State {$FES\leftarrow FES + \lambda$, $t\leftarrow t + 1$;}
    \EndWhile 
     \label{HECO-While-End}
    \Ensure the best individual $\mathbf{x} \in P$. 
\end{algorithmic}  

For the sake of self-contained presentation, we explain HECO-PDE in detail. Lines~\ref{HECO-Initialisation1} and~\ref{HECO-Initialisation2} are to  initialise several algorithm parameters, including:
\begin{enumerate}
    \item $\mu_0$ is the required initial population size  and   $\mu_{T_{\max}}$ is final size;
     \item  $FES_{\max}$ is a pre-defined maximum number of fitness evaluations;
     \item  $F$ and $CR$ are two circle memories for parameters;
     \item   $H$ is the size of historical memories;
     \item  $q_k, k=1, \cdots, 4$ are the initial probability distribution  of four strategies
     \item  $A$ is external archive; 
     \item $t$ is the number of generations. 
\end{enumerate}

{Lines~\ref{HECO-Generate-Population}-\ref{HECO-FES}} are to generate an initial population $P$ at random and evaluate the objective value $f(\mathbf{x})$  and constraint violation $v(\mathbf{x})$ for each individual $\mathbf{x}$. The number of fitness evaluation $FES=\mu_0$.
 
The while loop in  {lines~\ref{HECO-While-Start}-\ref{HECO-While-End}}  is the main part of the algorithm. 

In  {line~\ref{HECO-Weight}},  weights on functions $\tilde{e}, v, f$ are adjusted by the following formulas:  
\begin{align}
\begin{array}{lll}
    w_{1i,t} &=   \left(\frac{t}{T_{\max}}\right)^{20\times i}. \\
    w_{2i,t} &=  \frac{i}{\lambda}  \left( \frac{t}{T_{\max}}\right)^{\frac{5\times i}{\lambda}}.
\\
    w_{3i,t} &= \left(1-\frac{i}{\lambda}\right) \left(1 - \frac{t}{T_{\max}}\right)^{\frac{5\times i}{\lambda}}.
\end{array}
\end{align} 
Weights $w_{1i,t}$ and $w_{2i,t}$    increase  over  $t$ but $w_{3i,t}$ decreases over   $t$.

In  {line~\ref{HECO-Stratgies}},  sets $S_{F}$ and $S_{{CR}}$ are assigned to an empty set for each strategy. Here a strategy means a combination of a mutation operator and a crossover operator. 
 Sets $S_{F}$ and $S_{{CR}}$ are used to preserve successful values of $F$ in mutation and $CR$ in crossover  for each search strategy respectively. 

In  {line~\ref{HECO-Children}},  set $C$ (used for saving children population)  is also assigned to an empty set. 

In {line~\ref{HECO-population-Q}}, $\lambda$ individuals (denoted as  $Q$) are selected randomly from $P$ without putting them back to $P$.  Since  calculating the value of the equivalent function (\ref{equEquivalentFunction}) relies on ranking individuals in $Q$,  its size $\lambda$ is set to a small constant. Thus, the time complexity of ranking is a constant.  This is different from LSHADE44~\cite{polakova2017shade} which applies a search strategy to population $P$. The time complexity of ranking  in LSHADE44~\cite{polakova2017shade} is a function of $n$.


In lines~\ref{HECO-If-Start}-\ref{HECO-If-End},   a trial vector $y$ is generated by a mixed strategy.  PCA-projection is chosen with probability 0.1, or mutation and crossover are chosen with probability 0.9. $rand$ is a random number in $[0,1]$. 

In {lines~\ref{HECO-PCA1} and \ref{HECO-PCA2}},   PCA-projection is conducted on $Q$. Then $\lambda$ projected vectors are generated, denoted by $\mathbf{x}_{i}'$, $i=1, \cdots, \lambda$. Let $y_i =x'_i$.

In  {line~\ref{HECO-Stratgies2}},  the $k$th search  strategy is selected with probability $q_k$. A mechanism  of competition of strategies~\cite{tvrdik2006competitive,tvrdik2009adaptation} is employed to create trial points.  
The probability  $ q_k$ is adapted according to its success counts. A used strategy is considered successful if a generated trial point $y$ is better than the original point $x$. The selection probability 
\begin{align}
\label{equStrat}
    q_k = \frac{n_k + n_0}{\sum_{i=1}^{4}(n_i + n_0)},
\end{align}
where $n_k$ is the   count of the $k${th} strategy's successes, and $n_0>0$ is a constant. Parameters $F$ and $CR$   are generated from respective circle memories.

Then trial vector $\mathbf{y}_{i}$ is generated by
a strategy, i.e., mutation plus crossover.   There are two mutation operators used in the algorithm. The first one  is   current-to-pbest/1 mutation  as proposed in JADE~\cite{zhang2009jade}.
\begin{align}
    \label{equMutation1}
    \mathbf{u}_{i} = \mathbf{x}_{i} + F(\mathbf{x}_{pbest} - \mathbf{x}_{i}) + F(\mathbf{x}_{r_1} - \mathbf{x}_{r_2}),
\end{align} 
where $\mathbf{x}_{i}$ is the target point. $\mathbf{x}_{pbest}$ is randomly chosen from the top $100 p\%$ of population $P$ where $p\in (0, 1]$ a parameter. $\mathbf{x}_{r_1}$  is  uniformly chosen from population $P$. Individual $\mathbf{x}_{r_2}$ is randomly selected from $P\cup A$ where $A$ is an archive. $F$   is a mutation factor.

The second mutation is randrl/1  mutation~\cite{kaelo2006numerical}.  
\begin{align}
    \label{equMutation2}
    \mathbf{u}_{i} = \mathbf{x}_{r_1} + F(\mathbf{x}_{r_2} - \mathbf{x}_{r_3}),\\
    \label{equMutation3}
    \mathbf{u}_{i} = \mathbf{x}_{r^*_1} + F(\mathbf{x}_{r^*_2} - \mathbf{x}_{r^*_3}).
\end{align}
In mutation (\ref{equMutation2}),   $\mathbf{x}_{r_1}$, $\mathbf{x}_{r_2}$ and $\mathbf{x}_{r_3}$ are randomly chosen  from   population $P$. These three points are mutually distinct and different from $\mathbf{x}_{i}$.  
In mutation (\ref{equMutation3}),  $\mathbf{x}_{r_1}$, $\mathbf{x}_{r_2}$ and $\mathbf{x}_{r_3}$ are selected as mutation (\ref{equMutation2}) and then 
are ranked. $\mathbf{x}_{r^*_1}$ is the best individual while $\mathbf{x}_{r^*_2}$ and $\mathbf{x}_{r^*_3}$ denote the rest two. $F_k$ ($k=1,\cdots, K$) is the mutation factor. Kaelo and Ali~\cite{kaelo2006numerical} tested the randrl/1 mutation on a set of 50 difficult benchmark problems, with the results indicating that the use of this strategy speeds up the search by $30\%$ without significant decreasing the reliability as compared to searching with the original mutation operator.

There are two crossover operators used in the algorithm. The first one is binomial crossover: It combines coordinates of $\mathbf{x}_{i}$ with coordinates of mutant $\mathbf{u}_{i}$ into a trial point $\mathbf{y}_{i} $ 
\begin{align}
\label{equCRbio}
    y_{i,j} = 
    \left\{
    \begin{array}{ll}
    u_{i,j},     &\mbox{if} \quad rand_j(0, 1) \leq CR \mbox{ or }  j = j_{rand}, \\
    x_{i,j},     &\mbox{otherwise},
    \end{array}
    \right.
\end{align}
where the index $j_{rand}$ is a randomly chosen integer from $[1, D]$. $rand_j(0, 1)$ is   a uniform random number. $CR\in[0, 1]$ is a crossover control parameter.

The second crossover is the exponential crossover~\cite{storn1999system}. 
It combines the coordinates of   $\mathbf{x}_{i}$ with the coordinates of the mutant $\mathbf{u}_{i}$ into a trial point $\mathbf{y}_{i}$ as follows:
\begin{align}
\label{equCRexp}
   y_{i,j} = 
    \left\{
    \begin{array}{ll}
    u_{i,j},     &j = \langle l \rangle_D, \dots,\langle l+L-1 \rangle_D,\\
    x_{i,j},      &\mbox{otherwise},
   \end{array}
   \right.
\end{align}
where $\langle \rangle_D$ denotes a modulo function with modulus D. $l$ is the starting integer number randomly chosen from $[0,D - 1]$, and the integer $L$ is drawn from $[0, D - 1]$ subject to an exponential probability distribution.
Thus, four search strategies (combinations) can be produced. 

In {line~\ref{HECO-append}},   $\mathbf{y}_{i}$ is appended to  subpopulation $Q$, which results in an enlarged subpopulation $Q'$.  
In {line~\ref{HECO-normalize}}, the values of $\tilde{e}(\mathbf{x})$, $f(\mathbf{x})$ and $v(\mathbf{x})$ normalised by the max-min normalisation for each individual $\mathbf{x}$ in   $Q'$. Then the values of $f_i(\mathbf{y}_i)$ and $f_i(\mathbf{x}_i)$ are calculated according to formula~(\ref{equHEOP4}).
If $f_i(\mathbf{y}_{i}) < f_i(\mathbf{x}_{i})$, i.e., the trial point is better than its parent in terms of $f_i$, then in {line~\ref{HECO-update-archive}},  $\mathbf{x}_i$ is saved into  archive $A$, and  $\mathbf{y}_i$ into children set $C$. The $i$th individual in subpopulation $Q$ is used to minimise $f_i$. HECO-DE minimises $\lambda$ objective functions $f_i$  simultaneously. 
 
In {line~\ref{HECO-save-parameters}},  parameters $F$ in mutation and $CR$ in crossover are saved into respective sets $S_{F}$ and $S_{{CR}}$. Then the respective success counter is updated.

After completing search on subpopulation $Q$, in {lines~\ref{HECO-update-memory}},   circle memories $M_{F}$ and $M_{CR}$  are updated using respective $S_F$ and $S_{CR}$ for each strategy. The detail of this process refers to LSHADE44~\cite{polakova2017shade}.

In {line~\ref{HECO-update-P}},   population $P$ is updated by merging children $C$ and those individuals not involved in search. In {line~\ref{HECO-population-reduce1}-\ref{HECO-population-reduce2}}, if population size $|P|$ is greater than the required size $\mu_{t}$, then  $|P|-\mu_{t}$ non-best individuals are removed from   $P$ at random where $N_t$ is  given by
\begin{align}
\label{equPopulationSize}
 \textstyle  \mu_{t} = round\left(\mu_0 - \frac{t}{T_{\max}} (\mu_0 - \mu_{T_{\max}})\right).
\end{align} 

Similarly in {lines~\ref{HECO-archive1}}-\ref{HECO-archive2},  if archive size $|A|$ is greater than the maximum size $|A|_{\max}$, then  $|A|-|A|_{\max}$  individuals are removed from   $A$ at random.
 

Except PCA-projection, the other part of HECO-PDE is the same as HECO-DE. This design aims to evaluate the effectiveness of PCA-project without the interface of other factors.

\section{Comparative Experiments and Results}
\label{section:4}

In order to demonstrate the effectiveness of PCA-projection, HECO-PDE and PMODE are tested on the IEEE CEC 2017 benchmark suite in constrained optimization competition~\cite{cec2017online} and compared with the state-of-art EAs participated in the CEC 2018 competition~\cite{cec2018online}.

\subsection{The CEC 2017 Benchmark Suit}

The IEEE CEC 2017 benchmark suit in constrained optimisation competition~\cite{cec2018online} consists of $28$ problems with the dimension $D=10, 30, 50, 100$, therefore, $4\times 28$ instances in total. This suit was adopted by both CEC 2017 and 2018 competitions~\cite{cec2017online}. The detail of these problems are listed in Table~\ref{table:cec2017}.

\begin{table}
\centering
\caption{Details of 28 test problems.   $I$ is the number of inequality constraints, $E$ is the number of equality constraints}
\label{table:cec2017}
\scalebox{0.6}{
\begin{tabular}{|c|c|c|c|}
\hline
\multirow{2}{*}{\begin{tabular}[c]{@{}c@{}}Problem\\ Search Range\end{tabular}}                                          & \multirow{2}{*}{Type of Objective}                                                & \multicolumn{2}{c|}{Number of Constraints}                                                                                                                       \\ \cline{3-4} 
                                                                               &                                                                                   & $E$                                                                          & $I$                                                                                   \\ \hline
\multirow{2}{*}{\begin{tabular}[c]{@{}c@{}}C01\\ {[}-100,100{]}$^D$\end{tabular}} & \multirow{2}{*}{Non Separable}                                                    & \multirow{2}{*}{0}                                                         & \multirow{2}{*}{\begin{tabular}[c]{@{}c@{}}1\\ Separable\end{tabular}}              \\
                                                                               &                                                                                   &                                                                            &                                                                                     \\ \hline
\multirow{2}{*}{\begin{tabular}[c]{@{}c@{}}C02\\ {[}-100,100{]}$^D$\end{tabular}} & \multirow{2}{*}{\begin{tabular}[c]{@{}c@{}}Non Separable,\\ Rotated\end{tabular}} & \multirow{2}{*}{0}                                                         & \multirow{2}{*}{\begin{tabular}[c]{@{}c@{}}1\\ Non Separable, Rotated\end{tabular}} \\
                                                                               &                                                                                   &                                                                            &                                                                                     \\ \hline
\multirow{2}{*}{\begin{tabular}[c]{@{}c@{}}C03\\ {[}-100,100{]}$^D$\end{tabular}} & \multirow{2}{*}{Non Separable}                                                    & \multirow{2}{*}{\begin{tabular}[c]{@{}c@{}}1\\ Separable\end{tabular}}     & \multirow{2}{*}{\begin{tabular}[c]{@{}c@{}}1\\ Separable\end{tabular}}              \\
                                                                               &                                                                                   &                                                                            &                                                                                     \\ \hline
\multirow{2}{*}{\begin{tabular}[c]{@{}c@{}}C04\\ {[}-10,10{]}$^D$\end{tabular}}   & \multirow{2}{*}{Separable}                                                        & \multirow{2}{*}{0}                                                         & \multirow{2}{*}{\begin{tabular}[c]{@{}c@{}}2\\ Separable\end{tabular}}              \\
                                                                               &                                                                                   &                                                                            &                                                                                     \\ \hline
\multirow{2}{*}{\begin{tabular}[c]{@{}c@{}}C05\\ {[}-10,10{]}$^D$\end{tabular}}   & \multirow{2}{*}{Non Separable}                                                    & \multirow{2}{*}{0}                                                         & \multirow{2}{*}{\begin{tabular}[c]{@{}c@{}}2\\ Non Separable, Rotated\end{tabular}} \\
                                                                               &                                                                                   &                                                                            &                                                                                     \\ \hline
\multirow{2}{*}{\begin{tabular}[c]{@{}c@{}}C06\\ {[}-20,20{]}$^D$\end{tabular}}   & \multirow{2}{*}{Separable}                                                        & \multirow{2}{*}{6}                                                         & \multirow{2}{*}{\begin{tabular}[c]{@{}c@{}}0\\ Separable\end{tabular}}              \\
                                                                               &                                                                                   &                                                                            &                                                                                     \\ \hline

\multirow{2}{*}{\begin{tabular}[c]{@{}c@{}}C07\\ {[}-50,50{]}$^D$\end{tabular}}   & \multirow{2}{*}{Separable}                                                        & \multirow{2}{*}{\begin{tabular}[c]{@{}c@{}}2\\ Separable\end{tabular}}     & \multirow{2}{*}{0}                                                                  \\
                                                                               &                                                                                   &                                                                            &                                                                                     \\ \hline
\multirow{2}{*}{\begin{tabular}[c]{@{}c@{}}C08\\ {[}-100,100{]}$^D$\end{tabular}} & \multirow{2}{*}{Separable}                                                        & \multirow{2}{*}{\begin{tabular}[c]{@{}c@{}}2\\ Non Separable\end{tabular}} & \multirow{2}{*}{0}                                                                  \\
                                                                               &                                                                                   &                                                                            &                                                                                     \\ \hline
\multirow{2}{*}{\begin{tabular}[c]{@{}c@{}}C09\\ {[}-10,10{]}$^D$\end{tabular}}   & \multirow{2}{*}{Separable}                                                        & \multirow{2}{*}{\begin{tabular}[c]{@{}c@{}}2\\ Non Separable\end{tabular}} & \multirow{2}{*}{0}                                                                  \\
                                                                               &                                                                                   &                                                                            &                                                                                     \\ \hline
\multirow{2}{*}{\begin{tabular}[c]{@{}c@{}}C10\\ {[}-100,100{]}$^D$\end{tabular}}   & \multirow{2}{*}{Separable}                                                        & \multirow{2}{*}{\begin{tabular}[c]{@{}c@{}}2\\ Non Separable\end{tabular}} & \multirow{2}{*}{0}                                                                  \\
                                                                               &                                                                                   &                                                                            &                                                                                     \\ \hline

\multirow{2}{*}{\begin{tabular}[c]{@{}c@{}}C11\\ {[}-100,100{]}$^D$\end{tabular}} & \multirow{2}{*}{Separable}                                                    & \multirow{2}{*}{\begin{tabular}[c]{@{}c@{}}1\\ Non Separable\end{tabular}}     & \multirow{2}{*}{\begin{tabular}[c]{@{}c@{}}1\\ Non Separable\end{tabular}}              \\
                                                                               &                                                                                   &                                                                            &                                                                                     \\ \hline

\multirow{2}{*}{\begin{tabular}[c]{@{}c@{}}C12\\ {[}-100,100{]}$^D$\end{tabular}} & \multirow{2}{*}{Separable}                                                        & \multirow{2}{*}{0}                                                         & \multirow{2}{*}{\begin{tabular}[c]{@{}c@{}}2\\ Separable\end{tabular}}              \\
                                                                               &                                                                                   &                                                                            &                                                                                     \\ \hline
\multirow{2}{*}{\begin{tabular}[c]{@{}c@{}}C13\\ {[}-100,100{]}$^D$\end{tabular}} & \multirow{2}{*}{Non Separable}                                                    & \multirow{2}{*}{0}                                                         & \multirow{2}{*}{\begin{tabular}[c]{@{}c@{}}3\\ Separable\end{tabular}}              \\
                                                                               &                                                                                   &                                                                            &                                                                                     \\ \hline
\multirow{2}{*}{\begin{tabular}[c]{@{}c@{}}C14\\ {[}-100,100{]}$^D$\end{tabular}} & \multirow{2}{*}{Non Separable}                                                    & \multirow{2}{*}{\begin{tabular}[c]{@{}c@{}}1\\ Separable\end{tabular}}     & \multirow{2}{*}{\begin{tabular}[c]{@{}c@{}}1\\ Separable\end{tabular}}              \\
                                                                               &                                                                                   &                                                                            &                                                                                     \\ \hline
\multirow{2}{*}{\begin{tabular}[c]{@{}c@{}}C15\\ {[}-100,100{]}$^D$\end{tabular}} & \multirow{2}{*}{Separable}                                                        & \multirow{2}{*}{1}                                                         & \multirow{2}{*}{1}                                                                  \\
                                                                               &                                                                                   &                                                                            &                                                                                     \\ \hline
\multirow{2}{*}{\begin{tabular}[c]{@{}c@{}}C16\\ {[}-100,100{]}$^D$\end{tabular}} & \multirow{2}{*}{Separable}                                                        & \multirow{2}{*}{\begin{tabular}[c]{@{}c@{}}1\\ Non Separable\end{tabular}} & \multirow{2}{*}{\begin{tabular}[c]{@{}c@{}}1\\ Separable\end{tabular}}              \\
                                                                               &                                                                                   &                                                                            &                                                                                     \\ \hline
\multirow{2}{*}{\begin{tabular}[c]{@{}c@{}}C17\\ {[}-100,100{]}$^D$\end{tabular}} & \multirow{2}{*}{Non Separable}                                                    & \multirow{2}{*}{\begin{tabular}[c]{@{}c@{}}1\\ Non Separable\end{tabular}} & \multirow{2}{*}{\begin{tabular}[c]{@{}c@{}}1\\ Separable\end{tabular}}              \\
                                                                               &                                                                                   &                                                                            &                                                                                     \\ \hline
\multirow{2}{*}{\begin{tabular}[c]{@{}c@{}}C18\\ {[}-100,100{]}$^D$\end{tabular}} & \multirow{2}{*}{Separable}                                                        & \multirow{2}{*}{1}                                                         & \multirow{2}{*}{2}                                                                  \\
                                                                               &                                                                                   &                                                                            &                                                                                     \\ \hline
\multirow{2}{*}{\begin{tabular}[c]{@{}c@{}}C19\\ {[}-50,50{]}$^D$\end{tabular}}   & \multirow{2}{*}{Separable}                                                        & \multirow{2}{*}{0}                                                         & \multirow{2}{*}{\begin{tabular}[c]{@{}c@{}}2\\ Non Separable\end{tabular}}          \\
                                                                               &                                                                                   &                                                                            &                                                                                     \\ \hline
\multirow{2}{*}{\begin{tabular}[c]{@{}c@{}}C20\\ {[}-100,100{]}$^D$\end{tabular}} & \multirow{2}{*}{Non Separable}                                                    & \multirow{2}{*}{0}                                                         & \multirow{2}{*}{2}                                                                  \\
                                                                               &                                                                                   &                                                                            &                                                                                     \\ \hline
\multirow{2}{*}{\begin{tabular}[c]{@{}c@{}}C21\\ {[}-100,100{]}$^D$\end{tabular}} & \multirow{2}{*}{Rotated}                                                          & \multirow{2}{*}{0}                                                         & \multirow{2}{*}{\begin{tabular}[c]{@{}c@{}}2\\ Rotated\end{tabular}}                \\
                                                                               &                                                                                   &                                                                            &                                                                                     \\ \hline
\multirow{2}{*}{\begin{tabular}[c]{@{}c@{}}C22\\ {[}-100,100{]}$^D$\end{tabular}} & \multirow{2}{*}{Rotated}                                                          & \multirow{2}{*}{0}                                                         & \multirow{2}{*}{\begin{tabular}[c]{@{}c@{}}3\\ Rotated\end{tabular}}                \\
                                                                               &                                                                                   &                                                                            &                                                                                     \\ \hline
\multirow{2}{*}{\begin{tabular}[c]{@{}c@{}}C23\\ {[}-100,100{]}$^D$\end{tabular}} & \multirow{2}{*}{Rotated}                                                          & \multirow{2}{*}{\begin{tabular}[c]{@{}c@{}}1\\ Rotated\end{tabular}}       & \multirow{2}{*}{\begin{tabular}[c]{@{}c@{}}1\\ Rotated\end{tabular}}                \\
                                                                               &                                                                                   &                                                                            &                                                                                     \\ \hline
\multirow{2}{*}{\begin{tabular}[c]{@{}c@{}}C24\\ {[}-100,100{]}$^D$\end{tabular}} & \multirow{2}{*}{Rotated}                                                          & \multirow{2}{*}{\begin{tabular}[c]{@{}c@{}}1\\ Rotated\end{tabular}}       & \multirow{2}{*}{\begin{tabular}[c]{@{}c@{}}1\\ Rotated\end{tabular}}                \\
                                                                               &                                                                                   &                                                                            &                                                                                     \\ \hline
\multirow{2}{*}{\begin{tabular}[c]{@{}c@{}}C25\\ {[}-100,100{]}$^D$\end{tabular}} & \multirow{2}{*}{Rotated}                                                          & \multirow{2}{*}{\begin{tabular}[c]{@{}c@{}}1\\ Rotated\end{tabular}}       & \multirow{2}{*}{\begin{tabular}[c]{@{}c@{}}1\\ Rotated\end{tabular}}                \\
                                                                               &                                                                                   &                                                                            &                                                                                     \\ \hline
\multirow{2}{*}{\begin{tabular}[c]{@{}c@{}}C26\\ {[}-100,100{]}$^D$\end{tabular}} & \multirow{2}{*}{Rotated}                                                          & \multirow{2}{*}{\begin{tabular}[c]{@{}c@{}}1\\ Rotated\end{tabular}}       & \multirow{2}{*}{\begin{tabular}[c]{@{}c@{}}1\\ Rotated\end{tabular}}                \\
                                                                               &                                                                                   &                                                                            &                                                                                     \\ \hline
\multirow{2}{*}{\begin{tabular}[c]{@{}c@{}}C27\\ {[}-100,100{]}$^D$\end{tabular}} & \multirow{2}{*}{Rotated}                                                          & \multirow{2}{*}{\begin{tabular}[c]{@{}c@{}}1\\ Rotated\end{tabular}}       & \multirow{2}{*}{\begin{tabular}[c]{@{}c@{}}2\\ Rotated\end{tabular}}                \\
                                                                               &                                                                                   &                                                                            &                                                                                     \\ \hline
\multirow{2}{*}{\begin{tabular}[c]{@{}c@{}}C28\\ {[}-50,50{]}$^D$\end{tabular}}   & \multirow{2}{*}{Rotated}                                                          & \multirow{2}{*}{0}                                                         & \multirow{2}{*}{\begin{tabular}[c]{@{}c@{}}2\\ Rotated\end{tabular}}                \\
                                                                               &                                                                                   &                                                                            &                                                                                     \\ \hline
\end{tabular}}
\end{table}

\subsection{Experimental Settings}
As suggested by the CEC 2018 competition~\cite{cec2018online}, the maximum number of function  evaluations  $FES_{\max}=20000 D$. For each algorithm,  25 independent runs  were taken on each problem and dimension respectively. 

The  parameters   of  PMODE are set as follows.  For a fair comparison, the setting is chosen  as the same as that used in CMODE  although fine-tuning parameters of PMODE may lead to better results.  
\begin{enumerate}
    \item population size $\mu$ is set as 180;
    \item scaling factor $F$ is randomly chosen between 0.5 and 0.6;
    \item PCA crossover controlling parameter $CR$ is randomly chosen between 0.9 and 0.95; 
    \item subpopulation size $\lambda = 8$;
    \item the interval  of executing infeasible solution replacement $k=22$.
\end{enumerate}

The  parameters of HECO-PDE are set as follows.  Although fine-tuning parameters of HECO-PDE may lead to better results, the setting is chosen as the same as that used in HECO-DE for a fair comparison.  
\begin{enumerate}
    \item the number of subproblem $Q$  $\lambda = 12$;
    \item in strategy competition, $n_0 = 2$, $K = 4$, $\delta =1/20$;
    \item the size of historical memories  $H = 5$;
    \item the initial and final required population sizes  $\mu_0 = 12\times D$, $\mu_{T_{\max}} = \lambda$.
\end{enumerate}

\subsection{Description of EAs under Comparison}
HECO-PDE and  PMODE are compared with all seven EAs participated in the CEC 2018 constrained optimisation competition~\cite{cec2018online} apart from HECO-DE and CMODE. All EAs from the CEC 2017 competition belong to the single-objective method for COPs. Thus, a latest decomposition-based multi-objective EA, DeCODE~\cite{wang2018decomposition}, is also taken in the comparison.
These algorithms are described as follows. 

\begin{enumerate}
    \item CAL-SHADE~\cite{zamuda2017adaptive}: Success-History based Adaptive Differential Evolution Algorithm including liner population size reduction, enhanced with adaptive constraint violation handling, i.e. adaptive $\epsilon$-constraint handling.
    \item LSHADE+IDE~\cite{tvrdik2017simple}: A simple framework for cooperation of two advanced adaptive DE variants. The search process is divided into two stages: (i) search feasible solutions via minimizing  the mean violation and stopped if a number of feasible solutions are found. (ii) minimize the function value until the stop condition is reached.
    \item LSHADE44~\cite{polakova2017shade}: Success-History based Adaptive Differential Evolution Algorithm including liner population size reduction, uses three different additional strategies compete, with Deb's superiority of feasibility rule.
    \item UDE~\cite{trivedi2017unified}: Uses three trial vector generation strategies and two parameter settings. At each generation, UDE divides the current population into two sub-populations. In the first population, UDE employs all the three trial vector generation strategies on each target vector. For another one, UDE employs strategy adaption from learning experience from evolution in first population.
    \item MA-ES~\cite{hellwig2018matrix}: Combines the Matrix Adaptation Evolution Strategy for unconstrained optimization with well-known constraint handling techniques. It handles box-constraints by reflecting exceeding components into the predefined box. Additional in-/equality constraints are dealt with by application of two constraint handling techniques: $\epsilon$-level ordering and a repair step that is based on gradient approximation.
    \item IUDE~\cite{Trivedi2018improved}: An improved version of UDE. Different from UDE, local search and duplication operators have been removed, it employs a combination of $\epsilon$-constraint handling technique and Deb's superiority of feasibility rule.
    \item LSHADE-IEpsilon~\cite{fan2018lshade44}: An improved $\epsilon$-constrained handling method (IEpsilon) for solving constrained single-objective optimization problems. The IEpsilon method adaptively adjusts the value of $\epsilon$ according to the proportion of feasible solutions in the current population. Furthermore, a new mutation operator DE/randr1*/1 is proposed.
    \item DeCODE~\cite{wang2018decomposition}: A recent decomposition-based EA made use of the weighted sum approach to decompose the transformed bi-objective problem into a number of scalar optimisation subproblems and then applied differential evolution to solve them. They designed a strategy of adjusting weights and a restart strategy to tackle COPs with complicated constraints.
\end{enumerate}

\subsection{The Rules for Ranking Algorithms}
Because the CEC 2017 competition aimed to compare performance of  a group of EAs together, a set of rules was provided for ranking all algorithms~\cite{cec2017online}. Our comparison follows the same rules, which are listed as below.

\begin{enumerate}
	\item The procedure for ranking algorithms based on mean values:
    \begin{enumerate}
    	\item Rank the algorithms based on feasibility rate;
        \item Then rank the algorithms according to the mean violation amounts;
        \item At last, rank the algorithms in terms of mean objective function value.
    \end{enumerate}
    \item The procedure for ranking the algorithms based on the median solutions:
    \begin{enumerate}
    	\item A feasible solution is better than an infeasible solution;
        \item Rank feasible solutions based on their objective function values;
        \item Rank infeasible solutions according to their constraint violation amounts.
    \end{enumerate}
\item 
Ranking all algorithms on multiple 
problems: for each problem, algorithms' ranks are determined in terms of the mean values and median solutions
at maximum allowed number of evaluations, respectively. The total rank value of an  algorithm is 
calculated as below:
\begin{equation}
    \textrm{Rank value} = \sum_{i = 1}^{28}rank_i\textrm{(using mean value)} + \sum_{i = 1}^{28}rank_i\textrm{(using median solution)}
\end{equation}
\end{enumerate}

\subsection{Comparative Experimental  Results}
Table~\ref{table:rank_allD2} summarises  ranks of all  algorithms on  four dimensions and total ranks. HECO-PDE and PMODE got  lower rank values than HECO-DE and CMODE, respectively. This result clearly demonstrates that HECO-DE and CMODE are improved by PCA-projection. 
Moreover, HECO-PDE got the lowest rank value  among all compared algorithms. This result means that HECO-PDE enhanced with PCA-projection is the best in terms of the overall performance. 

\begin{table}[ht]
\centering
\caption{Total ranks of CMODE, PMODE, HECO-DE, HECO-PDE, DeCODE and seven  EAs in  CEC2018 competition}
\label{table:rank_allD2}
\scalebox{0.8}{
\begin{tabular}{|lccccc|}
\hline
Algorithm/Dimension    & $10D$ & $30D$ & $50D$ & $100D$ & Total \\ \hline
CAL\_LSAHDE(2017)      & 418   & 398   & 428   & 435    & 1679  \\
LSHADE44+IDE(2017)     & 299   & 365   & 385   & 353    & 1402  \\
LSAHDE44(2017)         & 319   & 313   & 308   & 310    & 1250  \\
UDE(2017)              & 330   & 344   & 345   & 390    & 1409  \\
MA\_ES(2018)           & 266   & 240   & 243   & 246    & 995   \\
IUDE(2018)             & 193   & 226   & 224   & 292    & 935   \\
LSAHDE\_IEpsilon(2018) & 199   & 246   & 292   & 333    & 1070  \\
DeCODE                 & 237   & 276   & 277   & 296    & 1086  \\
CMODE                  & 443   & 618   & 628   & 631    & 2320  \\
PMODE                  & 425   & 610   & 630   & 626    & 2291  \\
HECO-DE                & 173   & 164   & 177   & 183    & 697   \\
HECO-PDE               & 155   & 138   & 152   & 177    & 622   \\ \hline
\end{tabular}}
\end{table}

The detailed rank values of all algorithms on mean values and median solutions on 28 test problems with the dimension of 10D, 30D, 50D and 100D are shown in Table~\ref{table:compare_mean_10D}-\ref{table:compare_median_100D}, respectively. 

Regarding the test functions with $10D$, rank values based on mean values and median solution on the 28 test functions are reported in Table~\ref{table:compare_mean_10D} and~\ref{table:compare_median_10D}, respectively. As shown in Table~\ref{table:compare_mean_10D}, in terms of mean values with $10D$, HECO-DE and CMODE got total rank values, with $81$ and $236$ respectively. By contrast, HECO-PDE got the lowest rank value, with $67$ and PMODE also got a lower rank value than CMODE, with $218$. As shown in Table~\ref{table:compare_median_10D}, in terms of median solutions with $10D$, HECO-DE and CMODE got total rank values, with $91$ and $207$ respectively. By contrast, HECO-PDE got the lowest rank value, with $87$ while PMODE got the same rank value with CMODE, with $207$.

\begin{table}[ht]
\centering
\caption{Ranks based on \textbf{mean values} on the 28 functions of \textbf{10 dimensions}} 
\label{table:compare_mean_10D}
\scalebox{0.6}{
\begin{tabular}{|lccccccccccccccccccccccccccccc|}
\hline
\multicolumn{1}{|c}{Problem} & 1 & 2 & 3  & 4  & 5  & 6  & 7  & 8  & 9  & 10 & 11 & 12 & 13 & 14 & 15 & 16 & 17 & 18 & 19 & 20 & 21 & 22 & 23 & 24 & 25 & 26 & 27 & 28 & Total \\ \hline
CAL\_LSAHDE(2017)            & 1 & 1 & 10 & 10 & 12 & 6  & 7  & 12 & 11 & 12 & 7  & 1  & 12 & 7  & 8  & 10 & 11 & 12 & 1  & 8  & 12 & 12 & 9  & 10 & 10 & 12 & 10 & 1  & 235   \\
LSHADE44+IDE(2017)           & 1 & 1 & 8  & 7  & 1  & 12 & 6  & 1  & 1  & 2  & 1  & 3  & 1  & 9  & 5  & 8  & 6  & 10 & 4  & 4  & 2  & 7  & 8  & 8  & 8  & 6  & 11 & 10 & 151   \\
LSAHDE44(2017)               & 1 & 1 & 9  & 5  & 1  & 11 & 5  & 1  & 10 & 2  & 2  & 10 & 1  & 8  & 9  & 9  & 7  & 9  & 2  & 1  & 4  & 9  & 7  & 7  & 9  & 7  & 9  & 11 & 167   \\
UDE(2017)                    & 1 & 1 & 7  & 8  & 11 & 7  & 4  & 1  & 8  & 2  & 10 & 1  & 11 & 6  & 6  & 7  & 8  & 8  & 9  & 10 & 8  & 11 & 6  & 4  & 7  & 9  & 8  & 7  & 186   \\
MA\_ES(2018)                 & 1 & 1 & 1  & 9  & 1  & 5  & 2  & 1  & 1  & 2  & 5  & 12 & 7  & 10 & 12 & 1  & 12 & 1  & 12 & 9  & 9  & 8  & 10 & 6  & 1  & 8  & 2  & 9  & 158   \\
IUDE(2018)                   & 1 & 1 & 5  & 3  & 1  & 1  & 8  & 1  & 1  & 2  & 6  & 7  & 1  & 3  & 4  & 1  & 5  & 7  & 4  & 7  & 3  & 10 & 1  & 4  & 1  & 4  & 7  & 6  & 105   \\
LSAHDE\_IEpsilon(2018)       & 1 & 1 & 6  & 4  & 1  & 8  & 3  & 1  & 1  & 2  & 3  & 5  & 1  & 2  & 7  & 1  & 4  & 2  & 3  & 6  & 5  & 1  & 3  & 9  & 1  & 5  & 1  & 12 & 99    \\
DeCODE                       & 1 & 1 & 1  & 6  & 1  & 1  & 1  & 11 & 8  & 1  & 9  & 4  & 10 & 1  & 1  & 6  & 3  & 11 & 10 & 5  & 1  & 1  & 5  & 1  & 6  & 3  & 12 & 4  & 125   \\
CMODE                        & 1 & 1 & 11 & 11 & 1  & 10 & 12 & 1  & 12 & 10 & 12 & 6  & 7  & 11 & 11 & 11 & 9  & 6  & 11 & 12 & 11 & 1  & 11 & 11 & 11 & 11 & 6  & 8  & 236   \\
PMODE                        & 1 & 1 & 12 & 12 & 1  & 9  & 10 & 1  & 1  & 10 & 11 & 11 & 7  & 12 & 10 & 12 & 10 & 5  & 4  & 11 & 10 & 1  & 12 & 12 & 12 & 10 & 5  & 5  & 218   \\
HECO-DE                      & 1 & 1 & 1  & 1  & 1  & 1  & 11 & 1  & 1  & 2  & 8  & 9  & 1  & 5  & 2  & 1  & 1  & 4  & 4  & 3  & 7  & 1  & 4  & 3  & 1  & 1  & 3  & 3  & 82    \\
HECO-PDE                     & 1 & 1 & 1  & 1  & 1  & 1  & 9  & 1  & 1  & 2  & 4  & 8  & 1  & 3  & 2  & 1  & 1  & 3  & 4  & 2  & 6  & 1  & 2  & 2  & 1  & 2  & 4  & 2  & 68    \\ \hline
\end{tabular}}
\end{table}

\begin{table}[ht]
\centering
\caption{Ranks based on \textbf{median solution} on the 28 functions of \textbf{10 dimensions}} 
\label{table:compare_median_10D}
\scalebox{0.6}{
\begin{tabular}{|lccccccccccccccccccccccccccccc|}
\hline
\multicolumn{1}{|c}{Problem} & 1 & 2 & 3  & 4  & 5 & 6  & 7  & 8  & 9  & 10 & 11 & 12 & 13 & 14 & 15 & 16 & 17 & 18 & 19 & 20 & 21 & 22 & 23 & 24 & 25 & 26 & 27 & 28 & Total \\ \hline
CAL\_LSAHDE(2017)            & 1 & 1 & 10 & 10 & 1 & 8  & 5  & 12 & 12 & 1  & 4  & 3  & 1  & 8  & 8  & 10 & 10 & 11 & 1  & 8  & 4  & 1  & 9  & 12 & 10 & 10 & 11 & 1  & 183   \\
LSHADE44+IDE(2017)           & 1 & 1 & 9  & 5  & 1 & 10 & 6  & 1  & 1  & 2  & 2  & 4  & 1  & 10 & 7  & 9  & 6  & 10 & 4  & 4  & 5  & 1  & 10 & 9  & 9  & 8  & 10 & 2  & 148   \\
LSAHDE44(2017)               & 1 & 1 & 8  & 7  & 1 & 9  & 7  & 1  & 1  & 2  & 4  & 12 & 1  & 9  & 10 & 8  & 7  & 8  & 2  & 1  & 1  & 1  & 8  & 8  & 8  & 6  & 9  & 11 & 152   \\
UDE(2017)                    & 1 & 1 & 6  & 8  & 1 & 7  & 4  & 1  & 10 & 2  & 10 & 1  & 1  & 1  & 5  & 7  & 9  & 9  & 11 & 10 & 1  & 1  & 1  & 5  & 7  & 9  & 8  & 7  & 144   \\
MA\_ES(2018)                 & 1 & 1 & 1  & 9  & 1 & 1  & 2  & 1  & 1  & 2  & 2  & 5  & 1  & 3  & 12 & 1  & 8  & 1  & 10 & 9  & 6  & 1  & 3  & 7  & 1  & 7  & 1  & 10 & 108   \\
IUDE(2018)                   & 1 & 1 & 1  & 1  & 1 & 1  & 8  & 1  & 1  & 2  & 1  & 8  & 1  & 3  & 5  & 1  & 1  & 6  & 4  & 7  & 7  & 1  & 4  & 5  & 1  & 1  & 7  & 7  & 88    \\
LSAHDE\_IEpsilon(2018)       & 1 & 1 & 7  & 4  & 1 & 1  & 3  & 1  & 1  & 2  & 6  & 6  & 1  & 3  & 4  & 1  & 5  & 2  & 3  & 6  & 11 & 1  & 6  & 3  & 1  & 5  & 2  & 12 & 100   \\
DeCODE                       & 1 & 1 & 1  & 6  & 1 & 1  & 1  & 1  & 10 & 2  & 9  & 1  & 1  & 1  & 1  & 6  & 4  & 12 & 12 & 5  & 1  & 1  & 1  & 1  & 6  & 4  & 12 & 9  & 112   \\
CMODE                        & 1 & 1 & 11 & 12 & 1 & 11 & 11 & 1  & 1  & 12 & 12 & 7  & 1  & 11 & 11 & 11 & 11 & 5  & 4  & 12 & 8  & 1  & 11 & 10 & 11 & 11 & 6  & 2  & 207   \\
PMODE                        & 1 & 1 & 12 & 11 & 1 & 12 & 12 & 1  & 1  & 2  & 11 & 9  & 1  & 12 & 9  & 12 & 12 & 7  & 4  & 11 & 10 & 1  & 12 & 11 & 12 & 12 & 5  & 2  & 207   \\
HECO-DE                      & 1 & 1 & 1  & 1  & 1 & 1  & 9  & 1  & 1  & 2  & 8  & 11 & 1  & 6  & 2  & 1  & 2  & 4  & 4  & 3  & 9  & 1  & 7  & 4  & 1  & 3  & 3  & 2  & 91    \\
HECO-PDE                     & 1 & 1 & 1  & 1  & 1 & 1  & 10 & 1  & 1  & 2  & 7  & 10 & 1  & 6  & 2  & 1  & 2  & 3  & 4  & 2  & 12 & 1  & 5  & 2  & 1  & 2  & 4  & 2  & 87    \\ \hline
\end{tabular}}
\end{table}

Regarding the test functions with $30D$, rank values based on mean values and median solution on the 28 test functions are reported in Table~\ref{table:compare_mean_30D} and~\ref{table:compare_median_30D}, respectively. As shown in Table~\ref{table:compare_mean_30D}, in terms of mean values with $30D$, HECO-DE and CMODE got total rank values, with $82$ and $307$ respectively. By contrast, HECO-PDE got the lowest rank value, with $68$ and PMODE got a slightly lower rank value than CMODE, with $305$. As shown in Table~\ref{table:compare_median_30D}, in terms of median solutions with $30D$, HECO-DE and CMODE got total rank values, with $83$ and $311$ respectively. By contrast, HECO-PDE got the lowest rank value, with $71$ and PMODE got lower rank value than CMODE, with $306$.

\begin{table}[ht]
\centering
\caption{Ranks based on \textbf{mean values} on the 28 functions of \textbf{30 dimensions}}
\label{table:compare_mean_30D}
\scalebox{0.6}{
\begin{tabular}{|lccccccccccccccccccccccccccccc|}
\hline
\multicolumn{1}{|c}{Problem} & 1  & 2  & 3  & 4  & 5  & 6  & 7  & 8  & 9  & 10 & 11 & 12 & 13 & 14 & 15 & 16 & 17 & 18 & 19 & 20 & 21 & 22 & 23 & 24 & 25 & 26 & 27 & 28 & Total \\ \hline
CAL\_LSAHDE(2017)            & 1  & 1  & 10 & 10 & 10 & 6  & 8  & 1  & 10 & 10 & 4  & 10 & 10 & 10 & 11 & 12 & 10 & 11 & 1  & 3  & 10 & 10 & 10 & 12 & 10 & 10 & 10 & 1  & 222   \\
LSHADE44+IDE(2017)           & 1  & 1  & 9  & 6  & 1  & 11 & 4  & 10 & 1  & 9  & 3  & 7  & 7  & 9  & 7  & 8  & 7  & 9  & 4  & 5  & 8  & 7  & 9  & 9  & 8  & 7  & 11 & 9  & 187   \\
LSAHDE44(2017)               & 1  & 1  & 8  & 4  & 1  & 10 & 5  & 2  & 1  & 1  & 2  & 5  & 6  & 8  & 8  & 9  & 6  & 6  & 2  & 1  & 7  & 8  & 7  & 8  & 9  & 6  & 9  & 10 & 151   \\
UDE(2017)                    & 1  & 1  & 5  & 9  & 7  & 4  & 2  & 9  & 7  & 8  & 8  & 8  & 8  & 6  & 6  & 7  & 8  & 7  & 9  & 9  & 4  & 6  & 5  & 6  & 6  & 8  & 8  & 6  & 178   \\
MA\_ES(2018)                 & 1  & 1  & 1  & 8  & 1  & 3  & 1  & 2  & 1  & 1  & 1  & 9  & 1  & 7  & 12 & 1  & 9  & 2  & 10 & 10 & 9  & 1  & 6  & 7  & 1  & 9  & 1  & 4  & 120   \\
IUDE(2018)                   & 1  & 1  & 6  & 5  & 1  & 7  & 7  & 2  & 7  & 1  & 5  & 2  & 5  & 1  & 5  & 4  & 3  & 5  & 7  & 8  & 6  & 5  & 4  & 3  & 5  & 2  & 5  & 7  & 120   \\
LSAHDE\_IEpsilon(2018)       & 1  & 1  & 7  & 1  & 1  & 9  & 3  & 2  & 1  & 1  & 9  & 6  & 9  & 5  & 3  & 6  & 5  & 1  & 3  & 2  & 3  & 9  & 2  & 5  & 7  & 5  & 2  & 12 & 121   \\
DeCODE                       & 1  & 1  & 1  & 7  & 9  & 5  & 6  & 8  & 9  & 1  & 12 & 1  & 1  & 2  & 4  & 5  & 4  & 12 & 8  & 4  & 5  & 4  & 8  & 4  & 4  & 1  & 12 & 5  & 144   \\
CMODE                        & 12 & 12 & 12 & 11 & 12 & 8  & 11 & 12 & 12 & 12 & 10 & 12 & 11 & 11 & 10 & 10 & 12 & 10 & 11 & 11 & 12 & 12 & 11 & 10 & 11 & 11 & 7  & 11 & 307   \\
PMODE                        & 11 & 11 & 11 & 12 & 11 & 12 & 12 & 11 & 11 & 11 & 11 & 11 & 12 & 12 & 9  & 11 & 11 & 8  & 12 & 12 & 11 & 11 & 12 & 11 & 12 & 12 & 6  & 8  & 305   \\
HECO-DE                      & 1  & 1  & 1  & 2  & 8  & 1  & 9  & 2  & 1  & 1  & 7  & 4  & 1  & 4  & 1  & 1  & 1  & 4  & 4  & 6  & 2  & 3  & 3  & 2  & 1  & 4  & 3  & 3  & 81    \\
HECO-PDE                     & 1  & 1  & 1  & 2  & 1  & 1  & 10 & 2  & 1  & 1  & 6  & 3  & 1  & 3  & 1  & 1  & 2  & 3  & 4  & 7  & 1  & 2  & 1  & 1  & 1  & 3  & 4  & 2  & 67    \\ \hline
\end{tabular}}
\end{table}

\begin{table}[ht]
\centering
\caption{Ranks based on \textbf{median solution} on the 28 functions of \textbf{30 dimensions}}
\label{table:compare_median_30D}
\scalebox{0.6}{
\begin{tabular}{|lccccccccccccccccccccccccccccc|}
\hline
\multicolumn{1}{|c}{Problem} & 1  & 2  & 3  & 4  & 5  & 6  & 7  & 8  & 9  & 10 & 11 & 12 & 13 & 14 & 15 & 16 & 17 & 18 & 19 & 20 & 21 & 22 & 23 & 24 & 25 & 26 & 27 & 28 & Total \\ \hline
CAL\_LSAHDE(2017)            & 1  & 1  & 10 & 10 & 1  & 6  & 8  & 1  & 1  & 1  & 4  & 8  & 10 & 7  & 9  & 12 & 10 & 9  & 1  & 3  & 3  & 10 & 6  & 12 & 12 & 10 & 9  & 1  & 176   \\
LSHADE44+IDE(2017)           & 1  & 1  & 9  & 2  & 1  & 10 & 6  & 10 & 2  & 10 & 3  & 2  & 1  & 10 & 7  & 8  & 8  & 11 & 4  & 5  & 9  & 7  & 9  & 9  & 8  & 5  & 11 & 9  & 178   \\
LSAHDE44(2017)               & 1  & 1  & 8  & 5  & 1  & 9  & 5  & 2  & 2  & 2  & 2  & 7  & 8  & 9  & 8  & 9  & 7  & 5  & 2  & 1  & 8  & 8  & 8  & 7  & 9  & 8  & 10 & 10 & 162   \\
UDE(2017)                    & 1  & 1  & 5  & 9  & 1  & 4  & 2  & 9  & 8  & 9  & 8  & 9  & 7  & 6  & 6  & 5  & 9  & 7  & 7  & 9  & 3  & 6  & 1  & 6  & 6  & 9  & 7  & 6  & 166   \\
MA\_ES(2018)                 & 1  & 1  & 1  & 8  & 1  & 3  & 1  & 2  & 2  & 2  & 1  & 10 & 1  & 8  & 12 & 1  & 6  & 1  & 10 & 10 & 10 & 1  & 7  & 8  & 1  & 6  & 1  & 4  & 120   \\
IUDE(2018)                   & 1  & 1  & 6  & 6  & 1  & 8  & 4  & 2  & 8  & 2  & 6  & 4  & 1  & 1  & 3  & 4  & 3  & 6  & 7  & 8  & 3  & 1  & 1  & 1  & 5  & 1  & 5  & 7  & 106   \\
LSAHDE\_IEpsilon(2018)       & 1  & 1  & 7  & 1  & 1  & 7  & 3  & 2  & 2  & 2  & 9  & 5  & 9  & 3  & 5  & 7  & 5  & 2  & 3  & 2  & 7  & 9  & 4  & 5  & 7  & 2  & 2  & 12 & 125   \\
DeCODE                       & 1  & 1  & 1  & 6  & 1  & 5  & 7  & 2  & 8  & 2  & 12 & 1  & 1  & 1  & 3  & 5  & 4  & 12 & 9  & 4  & 3  & 1  & 10 & 4  & 4  & 7  & 12 & 5  & 132   \\
CMODE                        & 12 & 11 & 12 & 11 & 12 & 11 & 11 & 12 & 12 & 11 & 11 & 12 & 11 & 11 & 11 & 10 & 12 & 10 & 11 & 11 & 12 & 12 & 11 & 10 & 11 & 11 & 8  & 11 & 311   \\
PMODE                        & 11 & 12 & 11 & 12 & 11 & 12 & 12 & 11 & 11 & 12 & 10 & 11 & 12 & 12 & 10 & 11 & 11 & 8  & 12 & 12 & 11 & 11 & 12 & 11 & 10 & 12 & 6  & 8  & 305   \\
HECO-DE                      & 1  & 1  & 1  & 2  & 1  & 1  & 9  & 2  & 2  & 2  & 7  & 6  & 1  & 4  & 1  & 1  & 2  & 4  & 4  & 6  & 2  & 5  & 5  & 3  & 1  & 4  & 3  & 2  & 83    \\
HECO-PDE                     & 1  & 1  & 1  & 2  & 1  & 1  & 10 & 2  & 2  & 2  & 5  & 3  & 1  & 4  & 1  & 1  & 1  & 3  & 4  & 7  & 1  & 1  & 3  & 2  & 1  & 3  & 4  & 3  & 71    \\ \hline
\end{tabular}}
\end{table}

Regarding the test functions with $50D$, rank values based on mean values and median solution on the 28 test functions are reported in Table~\ref{table:compare_mean_50D} and~\ref{table:compare_median_50D}, respectively. As shown in Table~\ref{table:compare_mean_50D}, in terms of mean values with $50D$, HECO-DE and CMODE got total rank values, with $90$ and $314$ respectively. By contrast, HECO-PDE got the lowest rank value, with $77$ and PMODE got a slightly lower rank value than CMODE, with $312$. As shown in Table~\ref{table:compare_median_50D}, in terms of median solutions with $50D$, HECO-DE and CMODE got total rank values, with $87$ and $314$ respectively. By contrast, HECO-PDE got the lowest rank value, with $75$ while PMODE got a higher rank value than CMODE, with $318$.

\begin{table}[ht]
\centering
\caption{Ranks based on \textbf{mean values} on the 28 functions of \textbf{50 dimensions}}
\label{table:compare_mean_50D}
\scalebox{0.6}{
\begin{tabular}{|lccccccccccccccccccccccccccccc|}
\hline
\multicolumn{1}{|c}{Problem} & 1  & 2  & 3  & 4  & 5  & 6  & 7  & 8  & 9  & 10 & 11 & 12 & 13 & 14 & 15 & 16 & 17 & 18 & 19 & 20 & 21 & 22 & 23 & 24 & 25 & 26 & 27 & 28 & Total \\ \hline
CAL\_LSAHDE(2017)            & 10 & 10 & 10 & 10 & 9  & 4  & 8  & 10 & 10 & 8  & 4  & 7  & 10 & 10 & 10 & 10 & 10 & 8  & 1  & 12 & 6  & 9  & 10 & 10 & 10 & 10 & 7  & 1  & 234   \\
LSHADE44+IDE(2017)           & 9  & 1  & 9  & 5  & 1  & 10 & 5  & 9  & 7  & 10 & 1  & 4  & 7  & 9  & 7  & 8  & 8  & 7  & 4  & 4  & 9  & 5  & 8  & 9  & 9  & 8  & 10 & 9  & 192   \\
LSAHDE44(2017)               & 1  & 1  & 8  & 3  & 1  & 9  & 4  & 1  & 3  & 1  & 2  & 9  & 6  & 8  & 8  & 9  & 7  & 6  & 3  & 1  & 10 & 6  & 6  & 8  & 8  & 7  & 6  & 8  & 150   \\
UDE(2017)                    & 1  & 1  & 5  & 9  & 10 & 3  & 2  & 8  & 1  & 9  & 5  & 6  & 9  & 6  & 6  & 6  & 9  & 11 & 9  & 7  & 3  & 8  & 5  & 6  & 6  & 9  & 11 & 5  & 176   \\
MA\_ES(2018)                 & 1  & 1  & 1  & 8  & 1  & 1  & 1  & 2  & 8  & 1  & 3  & 10 & 8  & 7  & 9  & 1  & 6  & 1  & 10 & 9  & 8  & 4  & 7  & 7  & 1  & 5  & 1  & 4  & 126   \\
IUDE(2018)                   & 1  & 1  & 6  & 7  & 1  & 11 & 7  & 4  & 1  & 1  & 7  & 3  & 5  & 1  & 3  & 5  & 3  & 5  & 8  & 8  & 4  & 3  & 4  & 3  & 5  & 1  & 5  & 7  & 120   \\
LSAHDE\_IEpsilon(2018)       & 1  & 1  & 7  & 4  & 8  & 8  & 3  & 7  & 6  & 7  & 8  & 8  & 4  & 4  & 5  & 7  & 5  & 2  & 7  & 2  & 5  & 10 & 2  & 5  & 7  & 6  & 2  & 10 & 151   \\
DeCODE                       & 1  & 1  & 1  & 6  & 1  & 2  & 6  & 3  & 9  & 6  & 12 & 5  & 1  & 2  & 4  & 4  & 4  & 12 & 2  & 3  & 7  & 2  & 9  & 4  & 4  & 4  & 12 & 6  & 133   \\
CMODE                        & 12 & 12 & 11 & 11 & 11 & 12 & 12 & 12 & 11 & 12 & 11 & 11 & 12 & 12 & 12 & 11 & 12 & 9  & 12 & 10 & 11 & 11 & 11 & 11 & 11 & 11 & 9  & 11 & 314   \\
PMODE                        & 11 & 11 & 12 & 12 & 12 & 7  & 11 & 11 & 12 & 11 & 10 & 12 & 11 & 11 & 11 & 12 & 11 & 10 & 11 & 11 & 12 & 12 & 12 & 12 & 12 & 12 & 8  & 12 & 312   \\
HECO-DE                      & 1  & 1  & 1  & 1  & 1  & 6  & 9  & 6  & 3  & 5  & 9  & 1  & 3  & 5  & 1  & 1  & 2  & 3  & 4  & 5  & 1  & 7  & 3  & 2  & 1  & 3  & 3  & 2  & 90    \\
HECO-PDE                     & 1  & 1  & 1  & 1  & 1  & 5  & 10 & 5  & 3  & 4  & 6  & 2  & 2  & 3  & 1  & 1  & 1  & 4  & 4  & 6  & 2  & 1  & 1  & 1  & 1  & 2  & 4  & 3  & 77    \\ \hline
\end{tabular}}
\end{table}

\begin{table}[ht]
\centering
\caption{Ranks based on \textbf{median solution} on the 28 functions of \textbf{50 dimensions}}
\label{table:compare_median_50D}
\scalebox{0.6}{
\begin{tabular}{|lccccccccccccccccccccccccccccc|}
\hline
\multicolumn{1}{|c}{Problem} & 1  & 2  & 3  & 4  & 5  & 6  & 7  & 8  & 9  & 10 & 11 & 12 & 13 & 14 & 15 & 16 & 17 & 18 & 19 & 20 & 21 & 22 & 23 & 24 & 25 & 26 & 27 & 28 & Total \\ \hline
CAL\_LSAHDE(2017)            & 1  & 1  & 10 & 10 & 1  & 6  & 8  & 10 & 7  & 8  & 4  & 7  & 10 & 6  & 12 & 10 & 10 & 9  & 1  & 3  & 7  & 9  & 6  & 10 & 10 & 10 & 7  & 1  & 194   \\
LSHADE44+IDE(2017)           & 1  & 1  & 9  & 2  & 1  & 9  & 5  & 9  & 9  & 10 & 3  & 3  & 8  & 10 & 7  & 9  & 8  & 8  & 5  & 5  & 9  & 7  & 9  & 9  & 9  & 8  & 11 & 9  & 193   \\
LSAHDE44(2017)               & 1  & 1  & 8  & 5  & 1  & 8  & 4  & 1  & 3  & 1  & 1  & 10 & 7  & 9  & 9  & 8  & 7  & 6  & 4  & 1  & 10 & 8  & 8  & 8  & 8  & 7  & 6  & 8  & 158   \\
UDE(2017)                    & 1  & 1  & 5  & 9  & 10 & 5  & 2  & 8  & 1  & 9  & 5  & 5  & 9  & 7  & 6  & 6  & 9  & 7  & 10 & 8  & 1  & 6  & 2  & 7  & 6  & 9  & 10 & 5  & 169   \\
MA\_ES(2018)                 & 1  & 1  & 1  & 8  & 1  & 3  & 1  & 2  & 8  & 1  & 2  & 9  & 1  & 8  & 8  & 1  & 5  & 1  & 9  & 10 & 8  & 5  & 7  & 6  & 1  & 4  & 1  & 4  & 117   \\
IUDE(2018)                   & 1  & 1  & 6  & 7  & 1  & 10 & 7  & 3  & 1  & 1  & 6  & 4  & 1  & 3  & 1  & 5  & 1  & 5  & 2  & 9  & 4  & 4  & 2  & 1  & 4  & 2  & 5  & 7  & 104   \\
LSAHDE\_IEpsilon(2018)       & 1  & 1  & 7  & 1  & 1  & 7  & 3  & 7  & 3  & 7  & 8  & 8  & 6  & 1  & 5  & 7  & 6  & 2  & 8  & 2  & 5  & 10 & 5  & 5  & 7  & 6  & 2  & 10 & 141   \\
DeCODE                       & 1  & 1  & 1  & 6  & 1  & 4  & 6  & 6  & 10 & 6  & 12 & 5  & 1  & 3  & 4  & 4  & 4  & 12 & 3  & 4  & 6  & 3  & 10 & 4  & 4  & 5  & 12 & 6  & 144   \\
CMODE                        & 12 & 12 & 12 & 11 & 11 & 12 & 12 & 12 & 11 & 11 & 11 & 12 & 12 & 11 & 10 & 11 & 11 & 10 & 12 & 11 & 11 & 11 & 12 & 11 & 11 & 11 & 8  & 12 & 314   \\
PMODE                        & 11 & 11 & 11 & 12 & 12 & 11 & 11 & 11 & 12 & 12 & 10 & 11 & 11 & 12 & 11 & 12 & 12 & 11 & 11 & 12 & 12 & 12 & 11 & 12 & 12 & 12 & 9  & 11 & 318   \\
HECO-DE                      & 1  & 1  & 1  & 2  & 1  & 2  & 9  & 4  & 3  & 4  & 9  & 2  & 1  & 5  & 2  & 1  & 3  & 4  & 5  & 6  & 3  & 2  & 4  & 3  & 1  & 3  & 3  & 2  & 87    \\
HECO-PDE                     & 1  & 1  & 1  & 2  & 1  & 1  & 10 & 5  & 3  & 4  & 7  & 1  & 1  & 2  & 2  & 1  & 2  & 3  & 5  & 7  & 2  & 1  & 1  & 2  & 1  & 1  & 4  & 3  & 75    \\ \hline
\end{tabular}}
\end{table}

Regarding the test functions with $100D$, rank values based on mean values and median solution on the 28 test functions are reported in Table~\ref{table:compare_mean_100D} and~\ref{table:compare_median_100D}, respectively. As shown in Table~\ref{table:compare_mean_100D}, in terms of mean values with $100D$, HECO-DE got highest total rank value with $87$ and CMODE got $315$ respectively. By contrast, HECO-PDE got higher rank value than HECO-DE  with $88$ while PMODE got a lower rank value than CMODE, with $305$. As shown in Table~\ref{table:compare_median_100D}, in terms of median solutions with $100D$, HECO-DE and CMODE got total rank values, with $96$ and $315$ respectively. By contrast, HECO-PDE got the lowest rank value  with $89$, while PMODE got a  higher rank value than CMODE  with $321$.

According to the CEC 2018 competition rules, the ranks of HECO-PDE and HECO-DE are on the top two on each dimension but CMODE and PMODE on the bottom two. This result confirms our guess that MOEAs based on non-dominance, such as PMODE and CMODE, may not perform as good as decomposition-based MOEAs, such as  HECO-PDE and HECO-DE for solving COPs.

\begin{table}[ht]
\centering
\caption{Ranks based on \textbf{mean values} on the 28 functions of \textbf{100 dimensions}}
\label{table:compare_mean_100D}
\scalebox{0.6}{
\begin{tabular}{|lccccccccccccccccccccccccccccc|}
\hline
\multicolumn{1}{|c}{Problem} & 1  & 2  & 3  & 4  & 5  & 6  & 7  & 8  & 9  & 10 & 11 & 12 & 13 & 14 & 15 & 16 & 17 & 18 & 19 & 20 & 21 & 22 & 23 & 24 & 25 & 26 & 27 & 28 & Total \\ \hline
CAL\_LSAHDE(2017)            & 9  & 9  & 11 & 10 & 6  & 3  & 9  & 10 & 8  & 9  & 5  & 5  & 9  & 10 & 10 & 10 & 10 & 11 & 1  & 12 & 10 & 7  & 10 & 10 & 10 & 10 & 9  & 1  & 234   \\
LSHADE44+IDE(2017)           & 10 & 10 & 8  & 4  & 4  & 11 & 4  & 6  & 2  & 7  & 1  & 1  & 6  & 9  & 5  & 8  & 8  & 7  & 3  & 4  & 5  & 6  & 6  & 7  & 9  & 8  & 10 & 7  & 176   \\
LSAHDE44(2017)               & 1  & 1  & 7  & 3  & 3  & 10 & 5  & 1  & 1  & 1  & 4  & 10 & 7  & 8  & 8  & 9  & 6  & 6  & 2  & 3  & 9  & 5  & 8  & 6  & 8  & 6  & 6  & 9  & 153   \\
UDE(2017)                    & 8  & 8  & 5  & 9  & 10 & 4  & 2  & 9  & 10 & 8  & 6  & 2  & 10 & 5  & 7  & 5  & 9  & 8  & 9  & 8  & 4  & 10 & 2  & 8  & 5  & 9  & 11 & 5  & 196   \\
MA\_ES(2018)                 & 1  & 1  & 1  & 7  & 9  & 1  & 1  & 2  & 9  & 4  & 2  & 8  & 4  & 7  & 9  & 1  & 5  & 1  & 10 & 9  & 8  & 2  & 7  & 4  & 1  & 5  & 3  & 4  & 126   \\
IUDE(2018)                   & 1  & 1  & 9  & 8  & 5  & 12 & 7  & 3  & 6  & 5  & 3  & 3  & 8  & 1  & 4  & 6  & 3  & 5  & 8  & 7  & 2  & 8  & 5  & 5  & 6  & 3  & 5  & 8  & 147   \\
LSAHDE\_IEpsilon(2018)       & 7  & 7  & 6  & 5  & 8  & 9  & 3  & 7  & 7  & 6  & 8  & 4  & 5  & 4  & 6  & 7  & 7  & 2  & 6  & 2  & 3  & 9  & 4  & 9  & 7  & 7  & 4  & 10 & 169   \\
DeCODE                       & 1  & 1  & 1  & 6  & 7  & 2  & 6  & 8  & 3  & 10 & 10 & 6  & 3  & 6  & 3  & 4  & 4  & 12 & 7  & 1  & 1  & 1  & 9  & 3  & 4  & 4  & 12 & 6  & 141   \\
CMODE                        & 12 & 12 & 10 & 12 & 12 & 8  & 12 & 11 & 12 & 12 & 11 & 11 & 12 & 11 & 12 & 12 & 11 & 10 & 12 & 11 & 12 & 12 & 11 & 11 & 12 & 12 & 8  & 11 & 315   \\
PMODE                        & 11 & 11 & 12 & 11 & 11 & 6  & 11 & 12 & 11 & 11 & 12 & 12 & 11 & 12 & 11 & 11 & 12 & 9  & 11 & 10 & 11 & 11 & 12 & 12 & 11 & 11 & 7  & 12 & 305   \\
HECO-DE                      & 1  & 1  & 1  & 1  & 1  & 7  & 8  & 4  & 3  & 2  & 9  & 7  & 1  & 3  & 1  & 1  & 1  & 3  & 3  & 5  & 7  & 3  & 3  & 2  & 3  & 2  & 2  & 2  & 87    \\
HECO-PDE                     & 1  & 1  & 1  & 2  & 1  & 5  & 10 & 5  & 3  & 3  & 7  & 9  & 2  & 2  & 1  & 1  & 2  & 4  & 3  & 6  & 6  & 4  & 1  & 1  & 2  & 1  & 1  & 3  & 88    \\ \hline
\end{tabular}}
\end{table}

\begin{table}[ht]
\centering
\caption{Ranks based on \textbf{median solution} on the 28 functions of \textbf{100 dimensions}}
\label{table:compare_median_100D}
\scalebox{0.6}{
\begin{tabular}{|lccccccccccccccccccccccccccccc|}
\hline
\multicolumn{1}{|c}{Problem} & 1  & 2  & 3  & 4  & 5  & 6  & 7  & 8  & 9  & 10 & 11 & 12 & 13 & 14 & 15 & 16 & 17 & 18 & 19 & 20 & 21 & 22 & 23 & 24 & 25 & 26 & 27 & 28 & Total \\ \hline
CAL\_LSAHDE(2017)            & 9  & 9  & 10 & 10 & 7  & 5  & 9  & 10 & 6  & 9  & 5  & 4  & 9  & 1  & 10 & 10 & 10 & 9  & 1  & 1  & 3  & 7  & 6  & 10 & 10 & 10 & 10 & 1  & 201   \\
LSHADE44+IDE(2017)           & 10 & 10 & 9  & 1  & 3  & 9  & 5  & 6  & 2  & 8  & 4  & 2  & 6  & 10 & 6  & 8  & 7  & 6  & 3  & 5  & 7  & 5  & 7  & 8  & 9  & 7  & 6  & 8  & 177   \\
LSAHDE44(2017)               & 1  & 1  & 8  & 4  & 4  & 8  & 6  & 1  & 1  & 1  & 2  & 10 & 7  & 9  & 9  & 9  & 5  & 7  & 2  & 4  & 8  & 4  & 9  & 9  & 8  & 4  & 7  & 9  & 157   \\
UDE(2017)                    & 8  & 8  & 5  & 9  & 10 & 4  & 2  & 9  & 8  & 10 & 6  & 1  & 10 & 6  & 8  & 5  & 9  & 8  & 9  & 9  & 5  & 10 & 3  & 6  & 5  & 8  & 8  & 5  & 194   \\
MA\_ES(2018)                 & 1  & 1  & 1  & 7  & 9  & 2  & 1  & 2  & 10 & 5  & 1  & 6  & 4  & 8  & 4  & 1  & 4  & 1  & 10 & 10 & 6  & 3  & 8  & 4  & 1  & 5  & 1  & 4  & 120   \\
IUDE(2018)                   & 1  & 1  & 7  & 8  & 5  & 10 & 4  & 3  & 9  & 4  & 3  & 3  & 8  & 2  & 7  & 6  & 3  & 5  & 7  & 8  & 2  & 8  & 5  & 5  & 6  & 3  & 5  & 7  & 145   \\
LSAHDE\_IEpsilon(2018)       & 7  & 7  & 6  & 5  & 8  & 6  & 3  & 7  & 7  & 6  & 8  & 5  & 5  & 5  & 4  & 7  & 6  & 2  & 6  & 3  & 4  & 9  & 4  & 7  & 7  & 6  & 4  & 10 & 164   \\
DeCODE                       & 1  & 1  & 1  & 6  & 6  & 3  & 7  & 8  & 3  & 7  & 12 & 7  & 3  & 7  & 3  & 4  & 8  & 12 & 8  & 2  & 1  & 1  & 10 & 3  & 4  & 9  & 12 & 6  & 155   \\
CMODE                        & 12 & 12 & 11 & 12 & 11 & 11 & 12 & 11 & 12 & 12 & 10 & 11 & 11 & 11 & 11 & 12 & 11 & 10 & 11 & 11 & 12 & 12 & 11 & 11 & 11 & 12 & 11 & 11 & 316   \\
PMODE                        & 11 & 11 & 12 & 11 & 12 & 12 & 11 & 12 & 11 & 11 & 11 & 12 & 12 & 12 & 12 & 11 & 12 & 11 & 12 & 12 & 11 & 11 & 12 & 12 & 12 & 11 & 9  & 12 & 321   \\
HECO-DE                      & 1  & 1  & 1  & 3  & 1  & 7  & 8  & 4  & 3  & 2  & 9  & 9  & 2  & 4  & 1  & 1  & 1  & 3  & 3  & 6  & 10 & 2  & 2  & 2  & 3  & 2  & 3  & 2  & 96    \\
HECO-PDE                     & 1  & 1  & 1  & 1  & 1  & 1  & 10 & 5  & 3  & 3  & 7  & 8  & 1  & 3  & 1  & 1  & 2  & 4  & 3  & 7  & 9  & 6  & 1  & 1  & 2  & 1  & 2  & 3  & 89    \\ \hline
\end{tabular}}
\end{table}

\section{Conclusions}
\label{section:5}
From an experimental observation, we find that  given a valley landscape, the maximal variance direction in a population can be regarded as the  valley direction. Based on this finding,  a new search operator, called   PCA-projection, is proposed, in which PCA is used to project points along the maximal variance direction. PCA-projection can be easily added into an existing MOEA through a mixed strategy. We design two MOEAs enhanced with PCA-projection, called HECO-PDE and PMODE, for evaluating the effectiveness of this new operator.  Experimental results show that an EA enhanced with PCA-projection performs better than its corresponding opponent without this operator. Furthermore,  HECO-PDE is ranked  first on all dimensions  when compared with the state-of-art single-objective EAs  from the IEEE CEC 2018 competition and another recent MOEA (DeCODE) for constrained optimisation. This study also reveals that   decomposition-based MOEAs, such as HECO-PDE and HECO-DE, are competitive with best single-objective and multi-objective EAs  in constrained optimisation, but MOEAs based on non-dominance, such as PMODE and CMODE, may not perform so well.
For the future work,   PCA-projection can be applied to other EAs.

\section*{References} 

\clearpage

\section{Supplement}
This supplement provides detailed experimental results of PMODE and HECO-PDE.  25 independent runs of PMODE and HECO-PDE are taken on each problem and dimension respectively. The maximum function  evaluations is set to $20000\times D$, where $D$ is the dimension of an optimization problem. 

\subsection{Detailed Experimental results of PMODE}

Tables~\ref{table:fvalues10D}$-$\ref{table:fvalues100D} gives the experimental results of PMODE in terms of the best, median, worst, mean, standard deviation and feasibility rate of  the function values  $10D$, $30D$, $50D$ and $100D$. $c$ is the number of violated  constraints at the median solution: the sequence of three numbers indicate the number 
 of violations (including inequality and equality) by more than 1.0, in the range
 $[0.01, 1.0]$ and in the range $[0.0001, 0.01]$ respectively. $\overline{v}$ denotes 
 the mean value of the violations of all constraints at the median solution. SR is the 
 feasibility rate of the solutions obtained in 25 runs. $\overline{vio}$ denotes the mean
 constraint violation value of all the solutions of 25 runs.

\begin{table}[ht]
\centering
\caption{Function Values of PMODE Achieved for 10D ($MaxFES = 20000 \times D$)}
\label{table:fvalues10D}
\begin{scriptsize}
\begin{tabular}{|cccccccc|}
\hline
problem          & C01          & C02          & C03          & C04          & C05         & C06         & C07          \\ \hline
best             & 2.14611e-24  & 9.11866e-25  & 9.41726e+03  & 3.88033e+01  & 9.91336e-22 & 7.39402e+01 & -1.60857e+02 \\
median           & 5.68959e-24  & 4.44967e-24  & 1.71079e+04  & 8.32015e+01  & 3.52520e-20 & 1.78727e+02 & -7.46900e+01 \\
c                & 0 0 0        & 0 0 0        & 0 1 0        & 0 0 0        & 0 0 0       & 2 3 0       & 2 0 0        \\
$\overline{v}$   & 0.00000e+00  & 0.00000e+00  & 1.16861e-02  & 0.00000e+00  & 0.00000e+00 & 5.24831e-01 & 7.57086e+01  \\
Mean             & 8.63981e-24  & 6.48690e-24  & 5.94558e+04  & 8.13511e+01  & 6.08489e-20 & 2.16101e+02 & -5.61896e+01 \\
Worst            & 3.43159e-23  & 1.80413e-23  & 4.46449e+04  & 1.02793e+02  & 2.59659e-19 & 1.35417e+03 & -9.13489e+01 \\
std              & 7.75626e-24  & 4.92035e-24  & 8.34239e+04  & 1.39084e+01  & 6.96775e-20 & 2.51834e+02 & 5.98942e+01  \\
SR               & 100          & 100          & 4            & 100          & 100         & 16          & 4            \\
$\overline{vio}$ & 0.00000e+00  & 0.00000e+00  & 1.65910e-02  & 0.00000e+00  & 0.00000e+00 & 5.21616e-01 & 7.85737e+01  \\ \hline
problem          & C08          & C09          & C10          & C11          & C12         & C13         & C14          \\ \hline
best             & -1.34840e-03 & -4.97525e-03 & -5.09646e-04 & -1.62228e+02 & 3.98790e+00 & 5.15504e-21 & 2.39559e+00  \\
median           & -1.34840e-03 & -4.97525e-03 & -5.09645e-04 & -3.70137e+02 & 3.98863e+00 & 6.58037e-20 & 3.39761e+00  \\
c                & 0 0 0        & 0 0 0        & 0 0 0        & 1 0 0        & 0 0 0       & 0 0 0       & 0 0 0        \\
$\overline{v}$   & 0.00000e+00  & 0.00000e+00  & 0.00000e+00  & 2.40353e+02  & 0.00000e+00 & 0.00000e+00 & 4.58844e-06  \\
Mean             & -1.34840e-03 & -4.97525e-03 & -5.09645e-04 & -3.55967e+02 & 4.83886e+00 & 1.59463e-01 & 3.45367e+00  \\
Worst            & -1.34840e-03 & -4.97525e-03 & -5.09637e-04 & -5.43096e+02 & 1.46065e+01 & 3.98658e+00 & 3.67062e+00  \\
std              & 3.25752e-10  & 0.00000e+00  & 2.05524e-09  & 9.43249e+01  & 2.87977e+00 & 7.81207e-01 & 3.43063e-01  \\
SR               & 100          & 100          & 100          & 0            & 100         & 100         & 48           \\
$\overline{vio}$ & 0.00000e+00  & 0.00000e+00  & 0.00000e+00  & 2.37912e+02  & 0.00000e+00 & 0.00000e+00 & 5.32261e-05  \\ \hline
problem          & C15          & C16          & C17          & C18          & C19         & C20         & C21          \\ \hline
best             & 5.49772e+00  & 5.18363e+01  & 9.97584e-01  & 1.00000e+01  & 0.00000e+00 & 1.47930e+00 & 3.98790e+00  \\
median           & 1.17811e+01  & 5.65481e+01  & 9.14289e-01  & 3.62209e+01  & 0.00000e+00 & 1.81463e+00 & 3.98881e+00  \\
c                & 0 0 1        & 0 0 1        & 1 1 0        & 0 1 0        & 1 0 0       & 0 0 0       & 0 0 0        \\
$\overline{v}$   & 5.51058e-05  & 1.46131e-04  & 5.50540e+00  & 1.29979e-01  & 6.63359e+03 & 0.00000e+00 & 0.00000e+00  \\
Mean             & 1.21578e+01  & 6.15751e+01  & 9.80767e-01  & 4.10083e+01  & 0.00000e+00 & 1.81955e+00 & 5.46111e+00  \\
Worst            & 8.63678e+00  & 5.65464e+01  & 1.00244e+00  & 5.17217e+01  & 0.00000e+00 & 2.17692e+00 & 2.27853e+01  \\
std              & 3.90569e+00  & 8.31210e+00  & 5.62635e-02  & 1.44737e+01  & 0.00000e+00 & 1.76447e-01 & 4.99452e+00  \\
SR               & 40           & 12           & 0            & 4            & 0           & 100         & 100          \\
$\overline{vio}$ & 3.24071e-04  & 1.99125e-04  & 5.50824e+00  & 2.93346e+00  & 6.63359e+03 & 0.00000e+00 & 0.00000e+00  \\ \hline
problem          & C22          & C23          & C24          & C25          & C26         & C27         & C28          \\ \hline
best             & 4.75166e-21  & 2.40531e+00  & 5.49772e+00  & 5.65486e+01  & 1.51776e-01 & 6.28153e+01 & 0.00000e+00  \\
median           & 1.04790e-19  & 3.81500e+00  & 1.49229e+01  & 4.39817e+01  & 1.00954e+00 & 1.37228e+01 & 0.00000e+00  \\
c                & 0 0 0        & 0 0 0        & 0 0 1        & 0 0 1        & 1 0 1       & 0 2 0       & 1 0 0        \\
$\overline{v}$   & 0.00000e+00  & 0.00000e+00  & 2.11608e-04  & 1.46534e-04  & 5.50411e+00 & 2.76359e-01 & 6.63359e+03  \\
Mean             & 2.88192e-19  & 3.29617e+00  & 1.37914e+01  & 6.26432e+01  & 9.08549e-01 & 2.33382e+01 & 1.62221e+00  \\
Worst            & 2.27974e-18  & 3.31445e+00  & 1.49199e+01  & 7.53947e+01  & 1.00543e+00 & 2.82342e+00 & 6.37173e+00  \\
std              & 5.20942e-19  & 4.03648e-01  & 3.65075e+00  & 1.19573e+01  & 2.12119e-01 & 2.20528e+01 & 2.92278e+00  \\
SR               & 100          & 52           & 20           & 4            & 0           & 0           & 0            \\
$\overline{vio}$ & 0.00000e+00  & 6.82870e-05  & 4.23089e-04  & 2.31590e-04  & 5.46639e+00 & 2.48248e-01 & 6.63622e+03  \\ \hline
\end{tabular}
\end{scriptsize}
\end{table}

\begin{table}[ht]
\centering
\caption{Function Values of PMODE Achieved for 30D ($MaxFES = 20000 \times D$)}
\label{table:fvalues30D}
\begin{scriptsize}
\begin{tabular}{|cccccccc|}
\hline
problem          & C01         & C02          & C03         & C04          & C05         & C06         & C07          \\ \hline
best             & 8.45246e+00 & 1.31645e+00  & 1.21533e+05 & 1.62387e+02  & 2.37034e+01 & 8.68151e+02 & -7.78047e+01 \\
median           & 6.22106e+01 & 6.14986e+01  & 4.46899e+05 & 3.66171e+02  & 1.22896e+02 & 2.23162e+03 & -1.15271e+02 \\
c                & 0 0 0       & 0 0 0        & 0 1 0       & 0 0 0        & 0 0 0       & 4 1 0       & 2 0 0        \\
$\overline{v}$   & 0.00000e+00 & 0.00000e+00  & 1.56053e-02 & 0.00000e+00  & 0.00000e+00 & 3.04630e+00 & 1.40266e+03  \\
Mean             & 1.01581e+02 & 8.79346e+01  & 5.24967e+05 & 3.25711e+02  & 1.76987e+02 & 2.75522e+03 & -6.32051e+01 \\
Worst            & 3.13217e+02 & 5.79893e+02  & 2.19709e+05 & 3.91920e+02  & 8.55390e+02 & 1.96412e+03 & -1.85538e+02 \\
std              & 8.75557e+01 & 1.15326e+02  & 4.36345e+05 & 7.78934e+01  & 1.76071e+02 & 1.34597e+03 & 8.25769e+01  \\
SR               & 100         & 100          & 8           & 100          & 100         & 0           & 0            \\
$\overline{vio}$ & 0.00000e+00 & 0.00000e+00  & 2.97523e-02 & 0.00000e+00  & 0.00000e+00 & 3.07096e+00 & 1.37837e+03  \\ \hline
problem          & C08         & C09          & C10         & C11          & C12         & C13         & C14          \\ \hline
best             & 4.30883e+00 & -2.66551e-03 & 2.93934e+00 & -1.95483e+03 & 2.58898e+02 & 7.91928e+05 & 1.16005e+01  \\
median           & 8.11294e+00 & 1.04331e+00  & 1.12307e+00 & -2.39334e+03 & 8.71198e+02 & 4.03949e+06 & 1.73876e+01  \\
c                & 2 0 0       & 0 0 0        & 2 0 0       & 1 0 0        & 1 0 0       & 1 0 0       & 2 0 0        \\
$\overline{v}$   & 2.27078e+02 & 0.00000e+00  & 9.07387e+04 & 6.74605e+02  & 3.93445e+02 & 2.83340e+02 & 2.44359e+03  \\
Mean             & 9.09124e+00 & 1.22155e+00  & 5.79186e+00 & -1.58495e+03 & 9.71512e+02 & 1.01176e+07 & 1.69531e+01  \\
Worst            & 1.59694e+01 & 5.12011e+00  & 8.62571e+00 & -1.61106e+03 & 3.21002e+03 & 6.20270e+07 & 1.95298e+01  \\
std              & 3.15098e+00 & 1.49632e+00  & 3.96077e+00 & 5.24220e+02  & 5.76453e+02 & 1.34472e+07 & 1.88152e+00  \\
SR               & 0           & 60           & 0           & 0            & 0           & 0           & 0            \\
$\overline{vio}$ & 2.63526e+02 & 4.79756e-02  & 1.04036e+05 & 8.29760e+02  & 4.46996e+02 & 3.57934e+02 & 2.53791e+03  \\ \hline
problem          & C15         & C16          & C17         & C18          & C19         & C20         & C21          \\ \hline
best             & 1.17809e+01 & 2.01062e+02  & 1.02982e+00 & 2.55074e+02  & 4.15257e+01 & 8.22588e+00 & 2.89044e+02  \\
median           & 1.49226e+01 & 2.08916e+02  & 1.15328e+00 & 8.36250e+02  & 4.02094e+01 & 9.25366e+00 & 7.53484e+02  \\
c                & 0 0 0       & 0 0 1        & 2 0 0       & 1 0 0        & 1 0 0       & 0 0 0       & 1 0 0        \\
$\overline{v}$   & 9.70150e-06 & 7.20124e-05  & 2.62061e+02 & 4.75943e+05  & 2.14626e+04 & 0.00000e+00 & 3.14773e+02  \\
Mean             & 1.81898e+01 & 2.11052e+02  & 1.17571e+00 & 9.55141e+02  & 5.60410e+01 & 9.23266e+00 & 7.69190e+02  \\
Worst            & 2.12045e+01 & 2.01059e+02  & 1.45716e+00 & 2.05000e+03  & 8.07226e+01 & 1.00181e+01 & 1.32907e+03  \\
std              & 2.73591e+00 & 1.23599e+01  & 1.00234e-01 & 4.97848e+02  & 1.26095e+01 & 4.38197e-01 & 2.77606e+02  \\
SR               & 48          & 16           & 0           & 0            & 0           & 100         & 0            \\
$\overline{vio}$ & 1.84991e-04 & 1.85297e-04  & 3.06938e+02 & 9.84241e+05  & 2.14638e+04 & 0.00000e+00 & 3.45572e+02  \\ \hline
problem          & C22         & C23          & C24         & C25          & C26         & C27         & C28          \\ \hline
best             & 2.62086e+06 & 1.90652e+01  & 1.49225e+01 & 1.90066e+02  & 1.21225e+00 & 9.75533e+02 & 6.22618e+01  \\
median           & 2.21334e+07 & 2.03930e+01  & 1.80597e+01 & 2.01061e+02  & 1.59385e+00 & 5.32941e+03 & 7.70427e+01  \\
c                & 2 0 0       & 2 0 0        & 0 0 1       & 0 0 1        & 2 0 0       & 1 0 0       & 1 0 0        \\
$\overline{v}$   & 6.24786e+02 & 1.24452e+04  & 3.08672e-03 & 1.07590e-04  & 1.14320e+03 & 4.62537e+05 & 2.14813e+04  \\
Mean             & 4.79527e+07 & 2.04905e+01  & 1.89402e+01 & 2.36413e+02  & 1.65233e+00 & 3.53610e+03 & 9.64144e+01  \\
Worst            & 3.09761e+08 & 2.13998e+01  & 2.61751e+01 & 3.56724e+02  & 2.35945e+00 & 7.53113e+03 & 1.69217e+02  \\
std              & 6.33025e+07 & 7.08163e-01  & 2.51595e+00 & 3.83776e+01  & 3.21242e-01 & 1.96683e+03 & 2.58908e+01  \\
SR               & 0           & 0            & 36          & 28           & 0           & 0           & 0            \\
$\overline{vio}$ & 7.67227e+02 & 1.37688e+04  & 1.47415e+02 & 2.75679e+02  & 1.26016e+03 & 9.43760e+05 & 2.14813e+04  \\ \hline
\end{tabular}
\end{scriptsize}
\end{table}

\begin{table}[ht]
\centering
\caption{Function Values of PMODE Achieved for 50D ($MaxFES = 20000 \times D$)}
\label{table:fvalues50D}
\begin{scriptsize}
\begin{tabular}{|cccccccc|}
\hline
problem          & C01         & C02         & C03         & C04          & C05         & C06         & C07          \\ \hline
best             & 2.28064e+03 & 1.98396e+03 & 1.93920e+05 & 3.55349e+02  & 5.97421e+03 & 1.75834e+03 & -5.36955e+02 \\
median           & 4.52311e+03 & 3.66519e+03 & 8.18781e+06 & 4.77077e+02  & 1.92933e+04 & 8.24012e+03 & -1.18883e+02 \\
c                & 0 0 0       & 0 0 0       & 0 1 0       & 0 0 0        & 0 0 0       & 3 1 0       & 2 0 0        \\
$\overline{v}$   & 0.00000e+00 & 0.00000e+00 & 2.11433e-02 & 0.00000e+00  & 0.00000e+00 & 3.67257e+00 & 2.89321e+03  \\
Mean             & 4.57826e+03 & 3.86686e+03 & 2.56010e+06 & 5.28980e+02  & 2.07483e+04 & 5.21220e+03 & -1.52570e+02 \\
Worst            & 6.58848e+03 & 5.80477e+03 & 1.80087e+06 & 6.71820e+02  & 6.05995e+04 & 4.84112e+03 & -2.33206e+02 \\
std              & 9.65902e+02 & 9.24684e+02 & 3.45818e+06 & 1.19169e+02  & 1.16951e+04 & 2.34983e+03 & 1.35106e+02  \\
SR               & 100         & 100         & 8           & 100          & 100         & 16          & 0            \\
$\overline{vio}$ & 0.00000e+00 & 0.00000e+00 & 4.21099e-02 & 0.00000e+00  & 0.00000e+00 & 3.45598e+00 & 2.82329e+03  \\ \hline
problem          & C08         & C09         & C10         & C11          & C12         & C13         & C14          \\ \hline
best             & 1.02717e+01 & 5.74202e+00 & 1.74030e+01 & -1.47463e+03 & 3.75936e+03 & 4.56714e+07 & 1.83443e+01  \\
median           & 1.47237e+01 & 6.01563e+00 & 2.35321e+01 & -1.04457e+03 & 6.65909e+03 & 1.81715e+08 & 1.96039e+01  \\
c                & 2 0 0       & 1 0 0       & 2 0 0       & 1 0 0        & 1 0 0       & 2 0 0       & 2 0 0        \\
$\overline{v}$   & 1.36576e+03 & 1.25530e+02 & 2.42791e+06 & 4.02216e+03  & 3.25068e+03 & 2.07518e+03 & 1.03922e+04  \\
Mean             & 1.67054e+01 & 7.26445e+00 & 1.77968e+01 & -1.28194e+03 & 7.21543e+03 & 1.91988e+08 & 1.96436e+01  \\
Worst            & 2.12763e+01 & 9.49550e+00 & 1.47180e+01 & -9.95573e+02 & 1.61630e+04 & 6.34024e+08 & 2.09103e+01  \\
std              & 3.86802e+00 & 1.35384e+00 & 5.56564e+00 & 3.36372e+02  & 2.58950e+03 & 1.28596e+08 & 6.79060e-01  \\
SR               & 0           & 0           & 0           & 0            & 0           & 0           & 0            \\
$\overline{vio}$ & 1.34029e+03 & 2.09972e+02 & 2.50252e+06 & 4.14513e+03  & 3.50114e+03 & 2.07567e+03 & 1.00248e+04  \\ \hline
problem          & C15         & C16         & C17         & C18          & C19         & C20         & C21          \\ \hline
best             & 1.80641e+01 & 4.08407e+02 & 1.81562e+00 & 4.67375e+03  & 1.25353e+02 & 1.62035e+01 & 3.46898e+03  \\
median           & 2.23511e+01 & 4.03140e+02 & 2.69752e+00 & 1.04115e+04  & 1.35750e+02 & 1.75255e+01 & 7.01829e+03  \\
c                & 2 0 0       & 1 1 0       & 2 0 0       & 2 0 0        & 1 0 0       & 0 0 0       & 1 0 0        \\
$\overline{v}$   & 7.95359e+02 & 5.63050e+02 & 3.32053e+03 & 5.38874e+07  & 3.63004e+04 & 0.00000e+00 & 3.40038e+03  \\
Mean             & 2.41366e+01 & 4.36228e+02 & 2.61806e+00 & 8.17305e+03  & 1.28788e+02 & 1.74999e+01 & 6.90454e+03  \\
Worst            & 2.81105e+01 & 6.32413e+02 & 3.56693e+00 & 9.97025e+03  & 1.40672e+02 & 1.87708e+01 & 1.08995e+04  \\
std              & 5.05616e+00 & 6.03201e+01 & 3.83810e-01 & 2.02373e+03  & 1.62059e+01 & 5.63727e-01 & 1.86784e+03  \\
SR               & 8           & 4           & 0           & 0            & 0           & 100         & 0            \\
$\overline{vio}$ & 9.94883e+02 & 8.85680e+02 & 3.16162e+03 & 5.42517e+07  & 3.62998e+04 & 0.00000e+00 & 3.34437e+03  \\ \hline
problem          & C22         & C23         & C24         & C25          & C26         & C27         & C28          \\ \hline
best             & 6.58082e+08 & 2.09123e+01 & 3.70265e+01 & 5.58189e+02  & 4.31932e+00 & 1.50354e+04 & 1.72173e+02  \\
median           & 2.09933e+09 & 2.14984e+01 & 4.47071e+01 & 7.44848e+02  & 6.25467e+00 & 2.78702e+04 & 2.02544e+02  \\
c                & 2 0 0       & 2 0 0       & 2 0 0       & 2 0 0        & 2 0 0       & 2 0 0       & 1 0 0        \\
$\overline{v}$   & 7.26657e+03 & 4.93000e+04 & 9.13401e+03 & 7.52269e+03  & 1.04348e+04 & 4.85357e+07 & 3.63275e+04  \\
Mean             & 2.02531e+09 & 2.13180e+01 & 4.55004e+01 & 7.72193e+02  & 6.51825e+00 & 2.64735e+04 & 2.12540e+02  \\
Worst            & 4.14934e+09 & 2.14182e+01 & 7.05414e+01 & 9.87128e+02  & 9.74742e+00 & 5.25566e+04 & 3.08671e+02  \\
std              & 7.91057e+08 & 2.26072e-01 & 9.58023e+00 & 1.16580e+02  & 1.46486e+00 & 7.94598e+03 & 3.13167e+01  \\
SR               & 0           & 0           & 0           & 0            & 0           & 0           & 0            \\
$\overline{vio}$ & 6.99710e+03 & 4.94889e+04 & 9.41546e+03 & 7.98474e+03  & 1.09620e+04 & 5.85456e+07 & 3.63287e+04  \\ \hline
\end{tabular}
\end{scriptsize}
\end{table}

\begin{table}[ht]
\centering
\caption{Function Values of PMODE Achieved for 100D ($MaxFES = 20000 \times D$)}
\label{table:fvalues100D}
\begin{scriptsize}
\begin{tabular}{|cccccccc|}
\hline
problem          & C01         & C02         & C03         & C04          & C05         & C06         & C07          \\ \hline
best             & 2.19725e+04 & 3.08252e+04 & 8.81317e+05 & 8.80889e+02  & 1.74982e+05 & 4.28539e+03 & -5.36330e+01 \\
median           & 3.48956e+04 & 4.21954e+04 & 1.64864e+07 & 1.05770e+03  & 3.53987e+05 & 1.18699e+04 & -3.22051e+02 \\
c                & 0 0 0       & 0 0 0       & 0 1 0       & 0 0 0        & 0 0 0       & 5 0 0       & 2 0 0        \\
$\overline{v}$   & 0.00000e+00 & 0.00000e+00 & 4.56214e-02 & 0.00000e+00  & 0.00000e+00 & 6.78371e+00 & 6.86213e+03  \\
Mean             & 3.48826e+04 & 4.22447e+04 & 4.43083e+06 & 1.07373e+03  & 3.44449e+05 & 1.10830e+04 & -1.49642e+02 \\
Worst            & 4.50649e+04 & 5.53684e+04 & 1.44167e+07 & 1.40462e+03  & 5.35239e+05 & 5.70569e+03 & -9.51640e+01 \\
std              & 6.46300e+03 & 6.58780e+03 & 3.88766e+06 & 1.34842e+02  & 8.99856e+04 & 4.10730e+03 & 1.92704e+02  \\
SR               & 100         & 100         & 8           & 100          & 100         & 8           & 0            \\
$\overline{vio}$ & 0.00000e+00 & 0.00000e+00 & 6.03512e-02 & 0.00000e+00  & 0.00000e+00 & 6.29700e+00 & 6.90888e+03  \\ \hline
problem          & C08         & C09         & C10         & C11          & C12         & C13         & C14          \\ \hline
best             & 2.07891e+01 & 1.02028e+01 & 3.48310e+01 & -1.42141e+03 & 3.13441e+04 & 1.56586e+09 & 2.07419e+01  \\
median           & 2.76848e+01 & 8.33249e+00 & 3.71183e+01 & -9.50141e+02 & 3.75117e+04 & 3.12784e+09 & 2.12499e+01  \\
c                & 2 0 0       & 1 0 0       & 2 0 0       & 1 0 0        & 1 0 0       & 1 0 0       & 2 0 0        \\
$\overline{v}$   & 7.53044e+03 & 6.19357e+03 & 4.52030e+07 & 1.53424e+04  & 1.84512e+04 & 1.19932e+04 & 5.14897e+04  \\
Mean             & 2.57686e+01 & 9.37371e+00 & 3.81451e+01 & -1.44764e+03 & 3.79194e+04 & 3.21154e+09 & 2.12221e+01  \\
Worst            & 3.55284e+01 & 9.85473e+00 & 3.92237e+01 & -1.87066e+03 & 4.90522e+04 & 5.25677e+09 & 2.12777e+01  \\
std              & 4.91663e+00 & 8.36433e-01 & 5.42529e+00 & 2.32202e+02  & 4.38064e+03 & 9.73355e+08 & 1.97632e-01  \\
SR               & 0           & 0           & 0           & 0            & 0           & 0           & 0            \\
$\overline{vio}$ & 7.58180e+03 & 6.67623e+03 & 4.67914e+07 & 1.58347e+04  & 1.86740e+04 & 1.22397e+04 & 5.20788e+04  \\ \hline
problem          & C15         & C16         & C17         & C18          & C19         & C20         & C21          \\ \hline
best             & 3.77923e+01 & 1.29718e+03 & 6.87595e+00 & 3.08885e+04  & 2.94193e+02 & 3.74545e+01 & 2.74753e+04  \\
median           & 3.78513e+01 & 1.49338e+03 & 1.01553e+01 & 3.87156e+04  & 3.56124e+02 & 3.90835e+01 & 3.62471e+04  \\
c                & 2 0 0       & 2 0 0       & 2 0 0       & 2 0 0        & 1 0 0       & 0 0 0       & 1 0 0        \\
$\overline{v}$   & 1.52068e+04 & 1.30303e+04 & 1.81611e+04 & 1.06212e+09  & 7.33963e+04 & 0.00000e+00 & 1.78372e+04  \\
Mean             & 4.32079e+01 & 1.50611e+03 & 9.95507e+00 & 4.00109e+04  & 3.50146e+02 & 3.90733e+01 & 3.58695e+04  \\
Worst            & 5.12716e+01 & 1.65646e+03 & 1.20672e+01 & 4.21590e+04  & 4.24377e+02 & 4.02646e+01 & 4.50014e+04  \\
std              & 3.66450e+00 & 1.00916e+02 & 1.27062e+00 & 3.97574e+03  & 2.95260e+01 & 7.19816e-01 & 4.72648e+03  \\
SR               & 0           & 0           & 0           & 0            & 0           & 100         & 0            \\
$\overline{vio}$ & 1.49949e+04 & 1.33886e+04 & 1.77606e+04 & 1.03625e+09  & 7.33968e+04 & 0.00000e+00 & 1.76425e+04  \\ \hline
problem          & C22         & C23         & C24         & C25          & C26         & C27         & C28          \\ \hline
best             & 1.61392e+10 & 2.14640e+01 & 6.86652e+01 & 2.11668e+03  & 2.04175e+01 & 9.99239e+04 & 3.98592e+02  \\
median           & 3.19578e+10 & 2.15640e+01 & 7.90501e+01 & 2.58110e+03  & 2.85646e+01 & 1.39773e+05 & 5.50789e+02  \\
c                & 2 0 0       & 2 0 0       & 2 0 0       & 2 0 0        & 2 0 0       & 2 0 0       & 1 0 0        \\
$\overline{v}$   & 3.76342e+04 & 1.88806e+05 & 5.61628e+04 & 4.98537e+04  & 5.49798e+04 & 1.06865e+09 & 7.34298e+04  \\
Mean             & 3.42906e+10 & 2.14957e+01 & 7.90906e+01 & 2.52591e+03  & 2.92427e+01 & 1.26691e+05 & 4.94682e+02  \\
Worst            & 6.23980e+10 & 2.15599e+01 & 8.25268e+01 & 3.05519e+03  & 3.75458e+01 & 1.25816e+05 & 6.46487e+02  \\
std              & 1.15359e+10 & 1.46944e-01 & 4.83272e+00 & 2.06318e+02  & 4.91052e+00 & 1.78593e+04 & 5.21990e+01  \\
SR               & 0           & 0           & 0           & 0            & 0           & 0           & 0            \\
$\overline{vio}$ & 3.82006e+04 & 1.85570e+05 & 5.70819e+04 & 4.89022e+04  & 5.63358e+04 & 1.09278e+09 & 7.34291e+04  \\ \hline
\end{tabular}
\end{scriptsize}
\end{table}

\subsection{Experimental Results of HECO-PDE}
Tables~\ref{table:HECO-fvalues10D}$-$\ref{table:HECO-fvalues100D} shows the experimental results of HECO-PDE in terms of  the best, median, worst, mean, standard deviation and feasibility rate of  the function values  $10D$, $30D$, $50D$ and $100D$. $c$ is the number of violated  constraints at the median solution: the sequence of three numbers indicate the number 
 of violations (including inequality and equality) by more than 1.0, in the range
 $[0.01, 1.0]$ and in the range $[0.0001, 0.01]$ respectively. $\overline{v}$ denotes 
 the mean value of the violations of all constraints at the median solution. SR is the 
 feasibility rate of the solutions obtained in 25 runs. $\overline{vio}$ denotes the mean
 constraint violation value of all the solutions of 25 runs.

\begin{table}[ht]
\centering
\caption{Function Values of HECO-PDE Achieved for 10D ($MaxFES = 20000 \times D$)}
\label{table:HECO-fvalues10D}
\begin{scriptsize}
\begin{tabular}{|cccccccc|}
\hline
problem          & C01          & C02          & C03          & C04          & C05         & C06         & C07          \\ \hline
best             & 0.00000e+00  & 0.00000e+00  & 0.00000e+00  & 0.00000e+00  & 0.00000e+00 & 0.00000e+00 & -2.58748e+02 \\
median           & 0.00000e+00  & 0.00000e+00  & 0.00000e+00  & 0.00000e+00  & 0.00000e+00 & 0.00000e+00 & -1.17310e+03 \\
c                & 0 0 0        & 0 0 0        & 0 0 0        & 0 0 0        & 0 0 0       & 0 0 0       & 0 0 2        \\
$\overline{v}$   & 0.00000e+00  & 0.00000e+00  & 0.00000e+00  & 0.00000e+00  & 0.00000e+00 & 0.00000e+00 & 1.91686e-03  \\
Mean             & 0.00000e+00  & 0.00000e+00  & 0.00000e+00  & 0.00000e+00  & 0.00000e+00 & 0.00000e+00 & -7.46594e+02 \\
Worst            & 0.00000e+00  & 0.00000e+00  & 0.00000e+00  & 0.00000e+00  & 0.00000e+00 & 0.00000e+00 & -4.48664e+02 \\
std              & 0.00000e+00  & 0.00000e+00  & 0.00000e+00  & 0.00000e+00  & 0.00000e+00 & 0.00000e+00 & 7.86557e+02  \\
SR               & 100          & 100          & 100          & 100          & 100         & 100         & 4            \\
$\overline{vio}$ & 0.00000e+00  & 0.00000e+00  & 0.00000e+00  & 0.00000e+00  & 0.00000e+00 & 0.00000e+00 & 2.31995e-03  \\ \hline
problem          & C08          & C09          & C10          & C11          & C12         & C13         & C14          \\ \hline
best             & -1.34840e-03 & -4.97525e-03 & -5.09647e-04 & -1.68819e-01 & 3.98790e+00 & 0.00000e+00 & 2.37633e+00  \\
median           & -1.34840e-03 & -4.97525e-03 & -5.09647e-04 & -1.68792e-01 & 3.98883e+00 & 0.00000e+00 & 2.37633e+00  \\
c                & 0 0 0        & 0 0 0        & 0 0 0        & 0 0 0        & 0 0 0       & 0 0 0       & 0 0 0        \\
$\overline{v}$   & 0.00000e+00  & 0.00000e+00  & 0.00000e+00  & 0.00000e+00  & 0.00000e+00 & 0.00000e+00 & 0.00000e+00  \\
Mean             & -1.34840e-03 & -4.97525e-03 & -5.09647e-04 & -1.68092e-01 & 3.98918e+00 & 0.00000e+00 & 2.37633e+00  \\
Worst            & -1.34840e-03 & -4.97525e-03 & -5.09647e-04 & -1.58782e-01 & 3.99096e+00 & 0.00000e+00 & 2.37633e+00  \\
std              & 3.48093e-16  & 0.00000e+00  & 3.31451e-15  & 2.15091e-03  & 1.08240e-03 & 0.00000e+00 & 0.00000e+00  \\
SR               & 100          & 100          & 100          & 100          & 100         & 100         & 100          \\
$\overline{vio}$ & 0.00000e+00  & 0.00000e+00  & 0.00000e+00  & 0.00000e+00  & 0.00000e+00 & 0.00000e+00 & 0.00000e+00  \\ \hline
problem          & C15          & C16          & C17          & C18          & C19         & C20         & C21          \\ \hline
best             & 2.35612e+00  & 0.00000e+00  & 1.08553e-02  & 1.00000e+01  & 0.00000e+00 & 1.04005e-01 & 3.98796e+00  \\
median           & 2.35612e+00  & 0.00000e+00  & 1.08553e-02  & 5.04203e+01  & 0.00000e+00 & 3.68418e-01 & 3.98991e+00  \\
c                & 0 0 0        & 0 0 0        & 1 0 0        & 0 0 0        & 1 0 0       & 0 0 0       & 0 0 0        \\
$\overline{v}$   & 0.00000e+00  & 0.00000e+00  & 4.50000e+00  & 0.00000e+00  & 6.63359e+03 & 0.00000e+00 & 0.00000e+00  \\
Mean             & 2.35612e+00  & 0.00000e+00  & 1.08553e-02  & 3.91026e+01  & 0.00000e+00 & 3.58433e-01 & 3.98967e+00  \\
Worst            & 2.35612e+00  & 0.00000e+00  & 1.08553e-02  & 5.04203e+01  & 0.00000e+00 & 5.20013e-01 & 3.99246e+00  \\
std              & 1.11995e-15  & 0.00000e+00  & 1.08780e-16  & 1.81487e+01  & 0.00000e+00 & 1.07653e-01 & 1.26111e-03  \\
SR               & 100          & 100          & 0            & 100          & 0           & 100         & 100          \\
$\overline{vio}$ & 0.00000e+00  & 0.00000e+00  & 4.50000e+00  & 0.00000e+00  & 6.63359e+03 & 0.00000e+00 & 0.00000e+00  \\ \hline
problem          & C22          & C23          & C24          & C25          & C26         & C27         & C28          \\ \hline
best             & 3.48642e-27  & 2.37633e+00  & 2.35612e+00  & 0.00000e+00  & 1.08553e-02 & 9.05515e+01 & 0.00000e+00  \\
median           & 3.48642e-27  & 2.37633e+00  & 2.35612e+00  & 0.00000e+00  & 1.08553e-02 & 9.57215e+01 & 0.00000e+00  \\
c                & 0 0 0        & 0 0 0        & 0 0 0        & 0 0 0        & 1 0 0       & 0 0 0       & 1 0 0        \\
$\overline{v}$   & 0.00000e+00  & 0.00000e+00  & 0.00000e+00  & 0.00000e+00  & 4.50000e+00 & 0.00000e+00 & 6.63359e+03  \\
Mean             & 3.49026e-27  & 2.37633e+00  & 2.35612e+00  & 0.00000e+00  & 7.37529e-02 & 9.40671e+01 & 0.00000e+00  \\
Worst            & 3.51844e-27  & 2.37633e+00  & 2.35612e+00  & 0.00000e+00  & 7.10694e-01 & 9.57217e+01 & 0.00000e+00  \\
std              & 1.04062e-29  & 5.21413e-07  & 5.18534e-08  & 0.00000e+00  & 1.75700e-01 & 2.41167e+00 & 0.00000e+00  \\
SR               & 100          & 100          & 100          & 100          & 0           & 100         & 0            \\
$\overline{vio}$ & 0.00000e+00  & 0.00000e+00  & 0.00000e+00  & 0.00000e+00  & 4.50000e+00 & 0.00000e+00 & 6.63359e+03  \\ \hline
\end{tabular} 
\end{scriptsize}
\end{table}

\begin{table}[ht]
\centering
\caption{Function Values of HECO-PDE Achieved for 30D ($MaxFES = 20000 \times D$)}
\label{table:HECO-fvalues30D}
\begin{scriptsize}
\begin{tabular}{|cccccccc|}
\hline
problem          & C01          & C02          & C03          & C04          & C05         & C06         & C07          \\ \hline
best             & 0.00000e+00  & 0.00000e+00  & 0.00000e+00  & 1.35728e+01  & 0.00000e+00 & 0.00000e+00 & -2.45791e+03 \\
median           & 9.33075e-30  & 1.17959e-29  & 3.49071e-29  & 1.35728e+01  & 0.00000e+00 & 0.00000e+00 & -2.68829e+03 \\
c                & 0 0 0        & 0 0 0        & 0 0 0        & 0 0 0        & 0 0 0       & 0 0 0       & 0 0 2        \\
$\overline{v}$   & 0.00000e+00  & 0.00000e+00  & 0.00000e+00  & 0.00000e+00  & 0.00000e+00 & 0.00000e+00 & 7.93332e-04  \\
Mean             & 1.59799e-29  & 1.96421e-29  & 3.95614e-29  & 1.35728e+01  & 0.00000e+00 & 0.00000e+00 & -1.94723e+03 \\
Worst            & 7.51020e-29  & 6.70162e-29  & 1.34895e-28  & 1.35728e+01  & 0.00000e+00 & 0.00000e+00 & -2.51299e+03 \\
std              & 2.11761e-29  & 1.97404e-29  & 3.35121e-29  & 1.28144e-14  & 0.00000e+00 & 0.00000e+00 & 9.19644e+02  \\
SR               & 100          & 100          & 100          & 100          & 100         & 100         & 0            \\
$\overline{vio}$ & 0.00000e+00  & 0.00000e+00  & 0.00000e+00  & 0.00000e+00  & 0.00000e+00 & 0.00000e+00 & 1.05592e-03  \\ \hline
problem          & C08          & C09          & C10          & C11          & C12         & C13         & C14          \\ \hline
best             & -2.83981e-04 & -2.66551e-03 & -1.02842e-04 & -9.24713e-01 & 3.98253e+00 & 0.00000e+00 & 1.40852e+00  \\
median           & -2.83981e-04 & -2.66551e-03 & -1.02842e-04 & -8.77334e-01 & 3.98257e+00 & 0.00000e+00 & 1.40852e+00  \\
c                & 0 0 0        & 0 0 0        & 0 0 0        & 0 0 0        & 0 0 0       & 0 0 0       & 0 0 0        \\
$\overline{v}$   & 0.00000e+00  & 0.00000e+00  & 0.00000e+00  & 0.00000e+00  & 0.00000e+00 & 0.00000e+00 & 0.00000e+00  \\
Mean             & -2.83981e-04 & -2.66551e-03 & -1.02842e-04 & -1.87293e+02 & 3.98269e+00 & 0.00000e+00 & 1.40852e+00  \\
Worst            & -2.83981e-04 & -2.66551e-03 & -1.02840e-04 & -1.25684e+03 & 3.98341e+00 & 0.00000e+00 & 1.40852e+00  \\
std              & 5.00769e-11  & 3.79326e-16  & 2.41501e-10  & 3.71804e+02  & 2.14922e-04 & 0.00000e+00 & 7.00761e-16  \\
SR               & 100          & 100          & 100          & 64           & 100         & 100         & 100          \\
$\overline{vio}$ & 0.00000e+00  & 0.00000e+00  & 0.00000e+00  & 1.26005e+01  & 0.00000e+00 & 0.00000e+00 & 0.00000e+00  \\ \hline
problem          & C15          & C16          & C17          & C18          & C19         & C20         & C21          \\ \hline
best             & 2.35612e+00  & 0.00000e+00  & 3.08555e-02  & 3.00000e+01  & 0.00000e+00 & 2.50134e+00 & 3.98253e+00  \\
median           & 2.35612e+00  & 0.00000e+00  & 1.44943e+00  & 5.34725e+01  & 0.00000e+00 & 2.84742e+00 & 3.98262e+00  \\
c                & 0 0 0        & 0 0 0        & 1 0 0        & 0 0 0        & 1 0 0       & 0 0 0       & 0 0 0        \\
$\overline{v}$   & 0.00000e+00  & 0.00000e+00  & 1.45000e+01  & 0.00000e+00  & 2.13749e+04 & 0.00000e+00 & 0.00000e+00  \\
Mean             & 2.35612e+00  & 0.00000e+00  & 6.51135e-01  & 4.85514e+01  & 0.00000e+00 & 2.87611e+00 & 3.98269e+00  \\
Worst            & 2.35612e+00  & 0.00000e+00  & 9.51999e-01  & 5.82173e+01  & 0.00000e+00 & 3.19426e+00 & 3.98327e+00  \\
std              & 1.13742e-15  & 0.00000e+00  & 4.82474e-01  & 8.31285e+00  & 0.00000e+00 & 1.88661e-01 & 1.94653e-04  \\
SR               & 100          & 100          & 0            & 100          & 0           & 100         & 100          \\
$\overline{vio}$ & 0.00000e+00  & 0.00000e+00  & 1.49400e+01  & 0.00000e+00  & 2.13749e+04 & 0.00000e+00 & 0.00000e+00  \\ \hline
problem          & C22          & C23          & C24          & C25          & C26         & C27         & C28          \\ \hline
best             & 3.59077e-14  & 1.40852e+00  & 2.35612e+00  & 0.00000e+00  & 6.11945e-01 & 2.08760e+02 & 0.00000e+00  \\
median           & 7.91069e-10  & 1.40852e+00  & 2.35612e+00  & 0.00000e+00  & 8.53762e-01 & 2.51342e+02 & 3.33028e-06  \\
c                & 0 0 0        & 0 0 0        & 0 0 0        & 0 0 0        & 1 0 0       & 0 0 0       & 1 0 0        \\
$\overline{v}$   & 0.00000e+00  & 0.00000e+00  & 0.00000e+00  & 0.00000e+00  & 1.55000e+01 & 0.00000e+00 & 2.13749e+04  \\
Mean             & 4.13459e-08  & 1.40853e+00  & 2.35612e+00  & 0.00000e+00  & 8.41942e-01 & 2.36013e+02 & 4.23646e-06  \\
Worst            & 7.52530e-07  & 1.40858e+00  & 2.35612e+00  & 0.00000e+00  & 1.01187e+00 & 2.51347e+02 & 1.77740e-05  \\
std              & 1.48613e-07  & 1.46320e-05  & 8.05949e-08  & 0.00000e+00  & 8.41683e-02 & 2.04397e+01 & 3.62229e-06  \\
SR               & 100          & 100          & 100          & 100          & 0           & 100         & 0            \\
$\overline{vio}$ & 0.00000e+00  & 0.00000e+00  & 0.00000e+00  & 0.00000e+00  & 1.55000e+01 & 0.00000e+00 & 2.13749e+04  \\ \hline
\end{tabular}
\end{scriptsize}
\end{table}

\begin{table}[ht]
\centering
\caption{Function Values of HECO-PDE Achieved for 50D ($MaxFES = 20000 \times D$)}
\label{table:HECO-fvalues50D}
\begin{scriptsize}
\begin{tabular}{|cccccccc|}
\hline
problem          & C01          & C02          & C03          & C04          & C05         & C06         & C07          \\ \hline
best             & 1.04672e-28  & 4.57539e-29  & 1.81832e-28  & 1.35728e+01  & 0.00000e+00 & 0.00000e+00 & -3.78512e+03 \\
median           & 2.08259e-28  & 2.17442e-28  & 3.30138e-28  & 1.35728e+01  & 9.12120e-30 & 1.41843e+02 & -8.31534e+02 \\
c                & 0 0 0        & 0 0 0        & 0 0 0        & 0 0 0        & 0 0 0       & 0 0 0       & 0 0 2        \\
$\overline{v}$   & 0.00000e+00  & 0.00000e+00  & 0.00000e+00  & 0.00000e+00  & 0.00000e+00 & 0.00000e+00 & 5.66334e-04  \\
Mean             & 2.51258e-28  & 2.17115e-28  & 4.12290e-28  & 1.35728e+01  & 4.52846e-29 & 1.16974e+02 & -2.79070e+03 \\
Worst            & 5.16815e-28  & 4.33787e-28  & 8.70114e-28  & 1.35728e+01  & 4.25245e-28 & 1.66179e+02 & -3.52267e+03 \\
std              & 1.21265e-28  & 8.04570e-29  & 1.98135e-28  & 2.10181e-15  & 8.97841e-29 & 9.83836e+01 & 9.79449e+02  \\
SR               & 100          & 100          & 100          & 100          & 100         & 76          & 12           \\
$\overline{vio}$ & 0.00000e+00  & 0.00000e+00  & 0.00000e+00  & 0.00000e+00  & 0.00000e+00 & 2.42773e-04 & 8.24621e-04  \\ \hline
problem          & C08          & C09          & C10          & C11          & C12         & C13         & C14          \\ \hline
best             & -1.34466e-04 & -2.03709e-03 & -4.82659e-05 & -3.11531e+02 & 3.98145e+00 & 0.00000e+00 & 1.09995e+00  \\
median           & -1.33456e-04 & -2.03709e-03 & -4.82628e-05 & -1.62884e+03 & 3.98150e+00 & 1.33101e-27 & 1.09995e+00  \\
c                & 0 0 0        & 0 0 0        & 0 0 0        & 1 0 0        & 0 0 0       & 0 0 0       & 0 0 0        \\
$\overline{v}$   & 0.00000e+00  & 0.00000e+00  & 0.00000e+00  & 3.29899e+01  & 0.00000e+00 & 0.00000e+00 & 0.00000e+00  \\
Mean             & -1.31402e-04 & -2.03709e-03 & -4.82606e-05 & -1.54196e+03 & 4.22857e+00 & 1.59465e-01 & 1.10073e+00  \\
Worst            & -1.12487e-04 & -2.03709e-03 & -4.82487e-05 & -2.67003e+03 & 7.06997e+00 & 3.98662e+00 & 1.11946e+00  \\
std              & 4.80757e-06  & 1.40141e-15  & 4.96661e-09  & 6.80331e+02  & 8.37725e-01 & 7.81216e-01 & 3.82241e-03  \\
SR               & 100          & 100          & 100          & 0            & 100         & 100         & 100          \\
$\overline{vio}$ & 0.00000e+00  & 0.00000e+00  & 0.00000e+00  & 3.55113e+01  & 0.00000e+00 & 0.00000e+00 & 0.00000e+00  \\ \hline
problem          & C15          & C16          & C17          & C18          & C19         & C20         & C21          \\ \hline
best             & 2.35612e+00  & 0.00000e+00  & 6.04826e-02  & 4.42174e+01  & 0.00000e+00 & 5.15997e+00 & 3.98145e+00  \\
median           & 2.35612e+00  & 0.00000e+00  & 5.45231e-01  & 4.61632e+01  & 0.00000e+00 & 6.24310e+00 & 3.98149e+00  \\
c                & 0 0 0        & 0 0 0        & 1 0 0        & 0 0 0        & 1 0 0       & 0 0 0       & 0 0 0        \\
$\overline{v}$   & 0.00000e+00  & 0.00000e+00  & 2.55000e+01  & 0.00000e+00  & 3.61162e+04 & 0.00000e+00 & 0.00000e+00  \\
Mean             & 2.35612e+00  & 0.00000e+00  & 6.17102e-01  & 4.76107e+01  & 0.00000e+00 & 6.10145e+00 & 4.10501e+00  \\
Worst            & 2.35612e+00  & 0.00000e+00  & 9.20881e-01  & 6.22173e+01  & 0.00000e+00 & 6.92786e+00 & 7.06871e+00  \\
std              & 1.55368e-15  & 0.00000e+00  & 3.32818e-01  & 3.65297e+00  & 0.00000e+00 & 4.34170e-01 & 6.04962e-01  \\
SR               & 100          & 100          & 0            & 100          & 0           & 100         & 100          \\
$\overline{vio}$ & 0.00000e+00  & 0.00000e+00  & 2.52200e+01  & 0.00000e+00  & 3.61162e+04 & 0.00000e+00 & 0.00000e+00  \\ \hline
problem          & C22          & C23          & C24          & C25          & C26         & C27         & C28          \\ \hline
best             & 5.13492e-01  & 1.09995e+00  & 2.35612e+00  & 0.00000e+00  & 8.01254e-01 & 2.47635e+02 & 1.65653e-05  \\
median           & 1.40918e+01  & 1.09996e+00  & 2.35612e+00  & 0.00000e+00  & 9.36281e-01 & 2.64205e+02 & 3.28074e+00  \\
c                & 0 0 0        & 0 0 0        & 0 0 0        & 0 0 0        & 1 0 0       & 0 0 0       & 1 0 0        \\
$\overline{v}$   & 0.00000e+00  & 0.00000e+00  & 0.00000e+00  & 0.00000e+00  & 2.55000e+01 & 0.00000e+00 & 3.61248e+04  \\
Mean             & 1.34654e+01  & 1.09997e+00  & 2.35612e+00  & 3.02025e-14  & 9.31995e-01 & 2.58242e+02 & 2.76288e+00  \\
Worst            & 1.60992e+01  & 1.10009e+00  & 2.35612e+00  & 7.55062e-13  & 1.01406e+00 & 2.64215e+02 & 7.12744e+00  \\
std              & 2.82924e+00  & 3.13374e-05  & 1.72250e-08  & 1.47961e-13  & 5.29458e-02 & 7.95311e+00 & 2.53458e+00  \\
SR               & 100          & 100          & 100          & 100          & 0           & 100         & 0            \\
$\overline{vio}$ & 0.00000e+00  & 0.00000e+00  & 0.00000e+00  & 0.00000e+00  & 2.55000e+01 & 0.00000e+00 & 3.61255e+04  \\ \hline
\end{tabular}
\end{scriptsize}
\end{table}

\begin{table}[ht]
\centering
\caption{Function Values of HECO-PDE Achieved for 100D ($MaxFES = 20000 \times D$)}
\label{table:HECO-fvalues100D}
\begin{scriptsize}
\begin{tabular}{|cccccccc|}
\hline
problem          & C01         & C02         & C03          & C04          & C05         & C06         & C07          \\ \hline
best             & 4.74860e-24 & 2.08968e-23 & 1.09781e-23  & 1.35728e+01  & 1.24693e-16 & 5.85856e+02 & -6.04093e+03 \\
median           & 2.96855e-22 & 7.65701e-22 & 2.63175e-22  & 1.35728e+01  & 5.60537e-15 & 9.10968e+02 & -5.12059e+03 \\
c                & 0 0 0       & 0 0 0       & 0 0 0        & 0 0 0        & 0 0 0       & 0 0 0       & 0 0 2        \\
$\overline{v}$   & 0.00000e+00 & 0.00000e+00 & 0.00000e+00  & 0.00000e+00  & 0.00000e+00 & 0.00000e+00 & 3.43542e-04  \\
Mean             & 7.92774e-22 & 6.12831e-21 & 1.04964e-21  & 1.36666e+01  & 1.13652e-13 & 8.17227e+02 & -4.19316e+03 \\
Worst            & 4.80812e-21 & 4.10046e-20 & 7.98385e-21  & 1.59192e+01  & 1.63788e-12 & 8.24776e+02 & -5.64890e+03 \\
std              & 1.10948e-21 & 1.11876e-20 & 1.84224e-21  & 4.59810e-01  & 3.33214e-13 & 1.19466e+02 & 1.20840e+03  \\
SR               & 100         & 100         & 100          & 100          & 100         & 64          & 20           \\
$\overline{vio}$ & 0.00000e+00 & 0.00000e+00 & 0.00000e+00  & 0.00000e+00  & 0.00000e+00 & 1.07363e-03 & 4.02313e-04  \\ \hline
problem          & C08         & C09         & C10          & C11          & C12         & C13         & C14          \\ \hline
best             & 9.89426e-05 & 0.00000e+00 & -1.69783e-05 & -6.28804e+03 & 1.88621e+01 & 3.32042e+01 & 7.84202e-01  \\
median           & 4.62686e-04 & 0.00000e+00 & -1.66940e-05 & -6.98033e+03 & 3.16345e+01 & 3.32046e+01 & 7.86865e-01  \\
c                & 0 0 0       & 0 0 0       & 0 0 0        & 1 0 0        & 0 0 0       & 0 0 0       & 0 0 0        \\
$\overline{v}$   & 0.00000e+00 & 0.00000e+00 & 0.00000e+00  & 8.12562e+01  & 0.00000e+00 & 0.00000e+00 & 0.00000e+00  \\
Mean             & 4.54336e-04 & 0.00000e+00 & -1.67005e-05 & -6.83288e+03 & 3.11306e+01 & 3.36826e+01 & 7.96313e-01  \\
Worst            & 9.65755e-04 & 0.00000e+00 & -1.62289e-05 & -6.23668e+03 & 3.18122e+01 & 3.71880e+01 & 8.38008e-01  \\
std              & 1.64588e-04 & 0.00000e+00 & 2.14329e-07  & 3.27748e+02  & 2.50488e+00 & 1.29435e+00 & 1.52042e-02  \\
SR               & 100         & 100         & 100          & 0            & 100         & 100         & 100          \\
$\overline{vio}$ & 0.00000e+00 & 0.00000e+00 & 0.00000e+00  & 9.43117e+01  & 0.00000e+00 & 0.00000e+00 & 0.00000e+00  \\ \hline
problem          & C15         & C16         & C17          & C18          & C19         & C20         & C21          \\ \hline
best             & 2.35612e+00 & 0.00000e+00 & 2.59387e-01  & 4.62791e+01  & 0.00000e+00 & 1.47876e+01 & 1.20688e+01  \\
median           & 2.35612e+00 & 0.00000e+00 & 8.79248e-01  & 1.30626e+02  & 0.00000e+00 & 1.58000e+01 & 3.15911e+01  \\
c                & 0 0 0       & 0 0 0       & 1 0 0        & 1 0 0        & 1 0 0       & 0 0 0       & 0 0 0        \\
$\overline{v}$   & 0.00000e+00 & 0.00000e+00 & 5.05000e+01  & 2.11765e+01  & 7.29695e+04 & 0.00000e+00 & 0.00000e+00  \\
Mean             & 2.35612e+00 & 0.00000e+00 & 8.01474e-01  & 8.51306e+01  & 1.68587e-09 & 1.58580e+01 & 3.03467e+01  \\
Worst            & 2.35612e+00 & 0.00000e+00 & 1.01669e+00  & 4.48688e+01  & 4.21468e-08 & 1.72009e+01 & 3.21128e+01  \\
std              & 1.37309e-15 & 0.00000e+00 & 2.12703e-01  & 4.01418e+01  & 8.25906e-09 & 6.38805e-01 & 4.49296e+00  \\
SR               & 100         & 100         & 0            & 32           & 0           & 100         & 100          \\
$\overline{vio}$ & 0.00000e+00 & 0.00000e+00 & 5.05000e+01  & 1.18429e+01  & 7.29695e+04 & 0.00000e+00 & 0.00000e+00  \\ \hline
problem          & C22         & C23         & C24          & C25          & C26         & C27         & C28          \\ \hline
best             & 7.82296e+01 & 7.84209e-01 & 2.35612e+00  & 1.57066e+00  & 1.00167e+00 & 2.84751e+02 & 2.38846e+01  \\
median           & 2.12380e+04 & 7.84225e-01 & 2.35612e+00  & 1.57066e+00  & 1.04543e+00 & 2.84829e+02 & 5.38774e+01  \\
c                & 1 0 0       & 0 0 0       & 0 0 0        & 0 0 0        & 1 0 0       & 0 0 0       & 1 0 0        \\
$\overline{v}$   & 1.54193e+01 & 0.00000e+00 & 0.00000e+00  & 0.00000e+00  & 5.05000e+01 & 0.00000e+00 & 7.30594e+04  \\
Mean             & 3.23001e+04 & 7.84396e-01 & 2.35612e+00  & 5.27774e+00  & 1.05214e+00 & 2.99755e+02 & 5.15351e+01  \\
Worst            & 8.20487e+02 & 7.87071e-01 & 2.35612e+00  & 3.29866e+01  & 1.09745e+00 & 3.18890e+02 & 9.74830e+01  \\
std              & 8.81380e+04 & 5.65603e-04 & 3.39107e-09  & 6.63285e+00  & 2.54681e-02 & 1.68980e+01 & 1.51348e+01  \\
SR               & 48          & 100         & 100          & 100          & 0           & 100         & 0            \\
$\overline{vio}$ & 3.65205e+01 & 0.00000e+00 & 0.00000e+00  & 0.00000e+00  & 5.05000e+01 & 0.00000e+00 & 7.30575e+04  \\ \hline
\end{tabular}
\end{scriptsize}
\end{table}

\end{document}